\documentclass[journal]{IEEEtran}

\usepackage{xcolor,soul,framed} 

\colorlet{shadecolor}{yellow}
\usepackage[pdftex]{graphicx}
\graphicspath{{../pdf/}{../jpeg/}}
\DeclareGraphicsExtensions{.pdf,.jpeg,.png}

\usepackage[cmex10]{amsmath}
\usepackage{array}
\usepackage{mdwmath}
\usepackage{mdwtab}
\usepackage{eqparbox}
\usepackage{url}
\usepackage{pifont}
\include{times}
\usepackage{epsfig}
\usepackage{graphicx}
\usepackage{amsmath}
\usepackage{circledsteps}

\usepackage[lined,ruled,commentsnumbered,linesnumbered]{algorithm2e}

\usepackage{amssymb}
\usepackage{tabu}
\usepackage{siunitx}
\usepackage{adjustbox}
\usepackage{tabularx}
\usepackage{subfigure}
\usepackage{array}

\begin{document}
\bstctlcite{IEEEexample:BSTcontrol}
    \title{DVHN: A Deep Hashing Framework for Large-scale Vehicle Re-identification}

 \author{\IEEEauthorblockN{Yongbiao Chen\IEEEauthorrefmark{1},
Sheng Zhang\IEEEauthorrefmark{2},
Fangxin Liu\IEEEauthorrefmark{1}, 
Chenggang Wu\IEEEauthorrefmark{1}, 
Kaicheng Guo\IEEEauthorrefmark{1}, and
Zhengwei Qi\IEEEauthorrefmark{1}} \\
\IEEEauthorblockA{\IEEEauthorrefmark{1}School of Electronic Information and Electrical Engineering,
Shanghai Jiao Tong University, Shanghai, China} \\
\IEEEauthorblockA{\IEEEauthorrefmark{2}University of Southern California, Los Angeles, USA}

}


\maketitle

\begin{abstract}
 Vehicle re-identification,~ which seeks to match query vehicle images with tremendous gallery images ,~has been gathering proliferating momentum.  Conventional methods generally perform re-identification tasks by representing vehicle images as real-valued feature vectors and then ranking the gallery images by computing the corresponding Euclidean distances. Despite achieving remarkable retrieval accuracy, these methods require tremendous memory and computation when the gallery set is large, making them inapplicable in real-world scenarios. \par
In light of this limitation, in this paper, we make the very first attempt to investigate the integration of deep hash learning with vehicle re-identification.  We propose a deep hash-based vehicle re-identification framework, dubbed DVHN, which substantially reduces memory usage and promotes retrieval efficiency while reserving nearest neighbor search accuracy. Concretely,~DVHN directly learns discrete compact binary hash codes for each image by jointly optimizing the feature learning network and the hash code generating module. Specifically, we directly constrain the output from the convolutional neural network to be discrete binary codes and ensure the learned binary codes are optimal for classification. To optimize the deep discrete hashing framework, we further propose an alternating minimization method for learning binary similarity-preserved hashing codes. Extensive experiments on two widely-studied vehicle re-identification datasets- \textbf{VehicleID} and \textbf{VeRi}-~have demonstrated the superiority of our method against the state-of-the-art deep hash methods. \textbf{DVHN} of $2048$ bits can achieve 13.94\% and 10.21\% accuracy improvement in terms of \textbf{mAP} and \textbf{Rank@1} for \textbf{VehicleID (800)} dataset. For \textbf{VeRi}, we achieve 35.45\% and 32.72\% performance gains for \textbf{Rank@1} and \textbf{mAP}, respectively. 
\end{abstract}


%
\IEEEpeerreviewmaketitle


\section{Introduction}
Vehicle re-identification (vehicle ReID)\cite{liu2016deeprela}\cite{liu2017beyond}\cite{yan2017exploiting}\cite{guo2018learning}\cite{lou2019embedding}\cite{meng2020parsing} has been receiving growing attention among the computer vision research community. It targets retrieving the corresponding vehicle images in the gallery set given a query image. The general re-identification framework consists of two module:  feature learning  and metric learning. 1) the feature learning module is responsible for extracting discriminative feature embedding from the vehicle image. 2) the distance metric learning module~\cite{chu2019vehicle} is for preserving the distances of original images in the embedding space. Previous vehicle re-identification methods have achieved pronounced performances on widely-studied research datasets. Nonetheless,~it is not feasible to apply these techniques directly into a real-world scenario where the gallery image set normally contains an astronomical amount of images. The main reasons are demonstrated as follows. First, since existing methods learns a real-
valued feature vector for each image, the memory storage cost could be exceedingly high when there exists a large number of images. For instance, storing a 2048-dimensional feature vector of data type `float64' takes up 16 kilobytes. For a gallery set of 10 million vehicle images, the total memory storage cost could be up to 150 gigabytes. On top of that, directly computing the similarity between two 2048-dimensional feature vectors is quite inefficient\cite{lin2015deep} and costly, making it undesirable when the query speed is a critical concern.  \par
Recently, substantial research efforts have been devoted to deep learning-based hash methods owing to their low storage cost and high retrieval efficiency. The goal of deep hash is to learn a hash function that embeds images into compact binary hash codes in the hamming space while preserving their similarity in the original space\cite{cakir2019hashing}\cite{cao2016correlation}\cite{cao2017hashnet}\cite{lin2015deep}. Since a 2048-bit hamming code only takes up 256 bytes in memory, the total storage for a dataset of 10 million images is less than 2.4 gigabytes, saving up to 147 gigabytes of memory compared to using real-valued feature vectors. Further, as stated in \cite{ong2016improved}, the computation of Hamming distance between binary hash codes can be accelerated by using the built-in CPU hardware instruction-\textbf{XOR}. Generally, the hamming distance computation could be completed with several machine instructions, significantly faster than computing its euclidean distance counterpart. \par

Captivated by the before-mentioned benefits, one may naturally ponder the possibility of directly applying the off-the-shelf deep hash techniques into addressing the large-scale vehicle ReID problem.~ Although researchers have successfully applied deep hashing in the image retrieval, the unique features of the ReID task make it  non-trivial to apply these methods directly, usually with notable performance drops. The degraded performance could be ascribed to the fact that general-purposed deep hash methods fail to learn robust and discriminative features of vehicle ReID datasets.  For instance, canonical deep hash methods\cite{cao2017hashnet}\cite{cao2018deep} adopt a convolutional neural network to learn features and a pairwise loss module to guide the generation of hamming code. In a scenario where there are thousands, even millions of vehicle ids, and where viewpoint variation for each vehicle identity is considerable, these off-the-shelf deep hash methods can only learn sub-optimal feature representation, leading to sub-optimal hash codes. \par
In this paper, we introduce a novel deep hash-based framework for efficient large-scale vehicle ReID, dubbed DVHN, which fully exploit the merits of deep hashing learning while preserving the retrieval accuracy in large-scale vehicle re-identification tasks. Specifically, it consists of four modules: \textcircled{1}  a random triplet selection module, \textcircled{2} a convolutional neural network-based feature extraction module, \textcircled{3} a hash code generation module that adopts metric learning to enforce the learned hash codes to preserve the visual similarity and dissimilarity in the original raw pixel space and \textcircled{4} a discrete \textit{Hamming} hashing learning module. Regarding the similarity-preserving metric learning module, concretely, we adopt a batch-hard triplet loss \cite{hermans2017defense} to pull close visually-similar image pairs and push away visually-disparate images in the hamming space. In addition, to further propel the learned hash codes to capture more discriminative features, we adopt \textbf{Identity loss} which is essentially a cross-entropy classification loss by regarding each vehicle identity as a different class. As for the discrete \textit{Hamming} hashing module, we constrain the  continuous feature into binary codes directly with a \textit{sign} activation function. Subsequently, we train another classifier on the binary codes to ensure the learned codes are optimal for classification. Since the \textit{sign} function is no-differentiable, we propose a novel optimization method to train the \textbf{DVHN} framework, dubbed \textbf{alternative optimization}. In the test stage,~for all the gallery and query images, we encode them into binary Hamming codes and rank the gallery images according to their corresponding hamming distances from the query image. \par
To sum up, we make the following contributions:
\begin{enumerate}
   \item We analyze the key obstacles impeding applying existing vehicle ReID methods into the real-world, large-scale vehicle re-identification setting. Inspired by recent advancements in the deep hash research community, we make the very first attempt to address the challenging large-scale vehicle ReID problem by learning binary hamming hash codes.
    \item We propose a novel framework that learns the compact hamming code for vehicle images in an end-to-end manner. We propose a novel discrete \textit{Hamming} hashing learning module, which learns the binary \textit{Hamming} codes while ensuring the binary codes are ideal for classification. To tackle the challenging discrete optimization problem, we propose a novel \textbf{alternative optimization} scheme to train the framework.
    \item We implement several state-of-the-art deep hash methods with pytorch, download the open-sourced methods, and thoroughly test their performances on the standard vehicle ReID datasets. We trained 72 models from scratch and provided valuable data for research in this literature. Through comprehensive comparisons, we have demonstrated the effectiveness and evidenced our superiority against the existing state-of-the-art methods. 
   
\end{enumerate}

\section{Related Works}
\subsection{Hash for content retrieval}
Recent years have seen a surge in attention around the theme of hashing for large-scale content retrieval~\cite{indyk1997locality}\cite{weiss2008spectral}\cite{gong2012iterative}\cite{heo2012spherical}\cite{jegou2010product}\cite{ge2013optimized}. According to the way they extract the features, existing hashing methods could be categorized into two groups: shallow hashing methods and deep learning based-hash methods. \par
Typical shallow methods are reliant upon the handcraft features to learn hash codes. A canonical example is  \cite{indyk1997locality}, which seeks to find a locality-sensitive hash family where the probability of hash collisions for similar objects is much higher than those dissimilar ones. Later, based on LSH, \cite{charikar2002similarity} further developed another variant of LSH (dubbed SIMHASH) for cosine similarities in Euclidean space. Though effective, LSH-based hash methods require long code lengths to generate satisfactory performance. In 2008,~\cite{weiss2008spectral}, proposes to learn compact hash codes by minimizing the weighted Hamming distance of image pairs. Though these handcrafted feature-based shallow methods achieved success to some extent, when applied to real data where dramatic appearance variation exists, they generally fail to capture the discriminative semantic information, leading to compromised performances. In light of this dilemma, a wealth of deep learning-based hash methods have been proposed~\cite{cao2017hashnet}\cite{fan20deep}\cite{li2015feature}\cite{zhu2016deep}
\cite{liu2016deep}.~\cite{xia2014supervised} introduces a two-stage hashing scheme which first decomposes the similarity matrix $S$ into a product of $H$ and $H^T$ where each row of it is the hash code for a specific training image. Then, in the later stage,  the convolutional neural network is trained with the supervision signals from the learned hash codes.~Later, \cite{lai2015simultaneous} proposes to learn the features and hash codes in an end-to-end manner. \cite{zhu2016deep} proposes a Bayesian learning framework adopting pairwise loss for similarity preserving. \cite{cao2018deep} further proposes to substitute the previous probability generation function for neural network output logits with a Cauchy distribution to penalize similar image pairs with hamming distances larger than a threshold. \cite{wu2019deep} introduced a deep incremental hash framework that alleviates the burden of retraining the model when the gallery image set increases. \cite{fan20deep} further introduces a deep polarized loss for the hamming code generation, obviating the need for an additional quantization loss. Despite the success, these methods are not tailored for the large-scale vehicle re-identification scenario and could only achieve sub-optimal performances, as is illustrated before. 
\begin{figure*}
    \centering
    \includegraphics[width=7.1in]{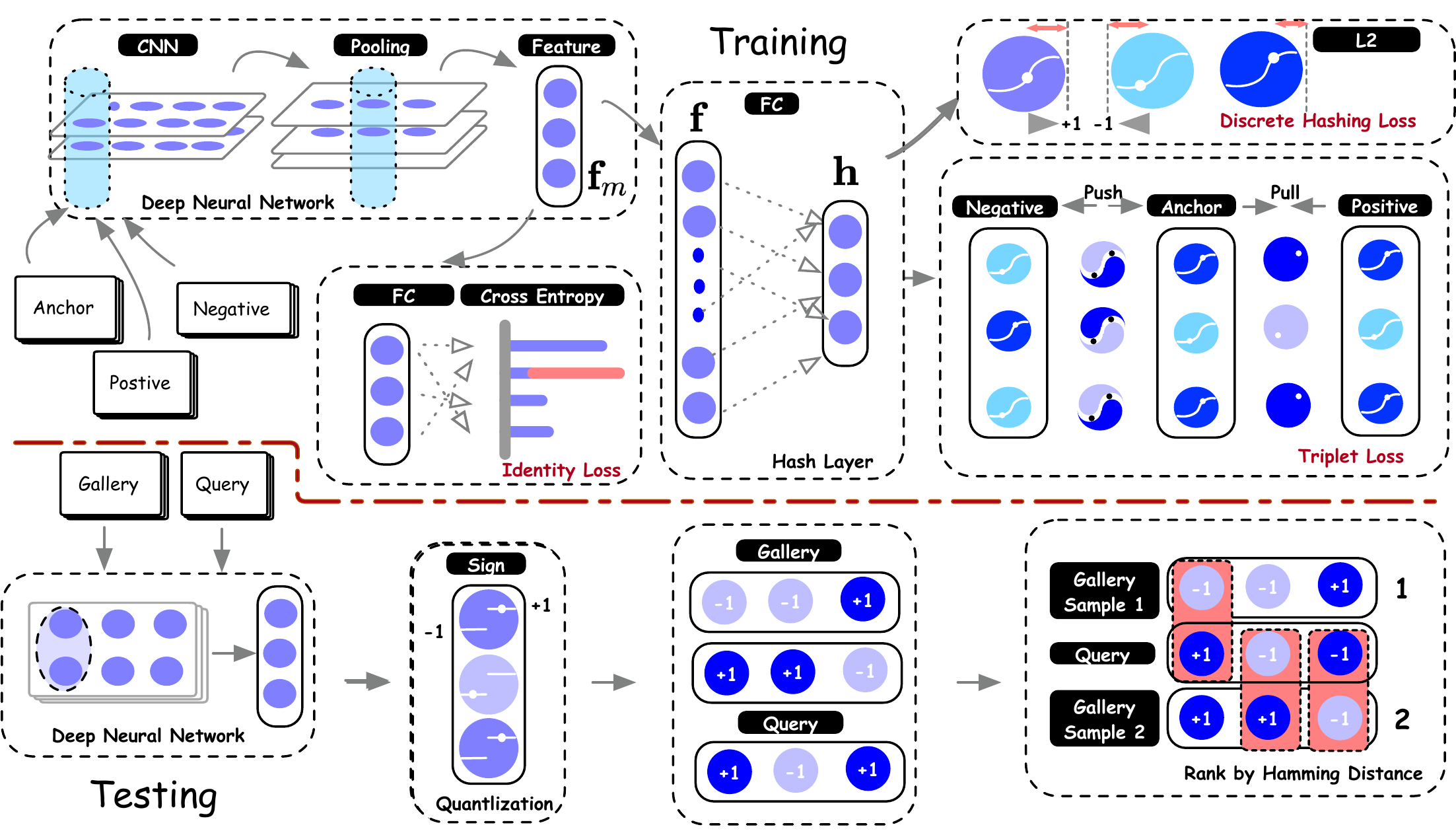}
    \caption{The framework of \textbf{DVHN}.~It basically contains two stages. The top part illustrates the process of learning to produce effective hash codes. The bottom demonstrates the testing process. During testing, the query and gallery images are encoded into binary discrete hash codes with the trained model. Then, the gallery images are ranked by computing their Hamming distances to the query image. }
     \label{fig:archi}
\end{figure*}

\subsection{Vehicle ReID} 
A wealth of vehicle ReID methods~\cite{liu2016large}\cite{liu2016deep}\cite{wang2017orientation} \cite{liu2017provid}\cite{meng2020parsing}\cite{baidisentangled} have been introduced in recent years given their extensive uses in digital surveillance, intelligent transportation~\cite{zhang2011data}, urban computing~\cite{zheng2014urban}. It is challenging considering its large intra-class variation and small inter-class differences. Some works strive to boost the discriminability of the learned representations. \cite{vehicleID} proposes a two-stream network to project vehicle images into comparable euclidean space. \cite{yan2017exploiting} proposes to learn feature representation by exploiting multi-grain ranking constraints.~\cite{guo2018learning} further proposes to learn robust features in a coarse-to-fine fashion. In 2019,~\cite{he2019part} adopts the vehicle part information to 
assist in the global feature learning. \cite{lou2019embedding} introduces a novel end-to-end adversarial learning-base network to learn robust and discriminative features. \cite{zheng2020vehiclenet} proposes to construct another large dataset named "VehicleNet" and innovated a two-stage learning framework to learn robust feature representation.
Another line of research focuses on learning viewpoint-invariant embedding for re-identification. Canonical examples include \cite{wang2017orientation}\cite{zhou2018aware}\cite{tang2019pamtri}\cite{baidisentangled}.\cite{wang2017orientation} proposes to mine the keypoint localization information to align and extract viewpoint-invariant information.~\cite{zhou2018aware} further introduces a scheme which utilizes single-view input information to generate multi-view features.~\cite{tang2019pamtri} innovates a pose-aware multi-task learning approach to learn viewpoint-invariant features. \cite{baidisentangled} introduces to disentangle the feature representation into orientation-specific and orientation-invariant features and, in the later stage, use the orientation-invariant feature to perform vehicle ReID task. \cite{zhu2019vehicle} proposes to extract quadruple directional deep features for vehicle-re identification while \cite{chen2020vehicle} proposes an end-to-end distance-based deep neural network combining multi-regional features to learn to distinguish local and global differences of a vehicle image. . \cite{liu2020beyond} explores vehicle parsing to learn discriminative part-level features an build a part-neighboring graph to model the correlation between every part features. To alleviate the influence of the drastic appearance variation, ~\cite{jin2021model} further introduces a novel multi-center metric learning framework which models the latent views from the vehicle visual appearance directly without the need for extra labels. 
Notwithstanding the successes in this literature, all the previously mentioned vehicle ReID methods are not applicable in the large-scale ReID settings where the database stores a significantly larger number of image items.

\section{Methodology}

In this paper, we denote mapping functions with calligraphic uppercase letters, such as $\mathcal{F}$. We denote sets with bold-face uppercase letters such as $\mathbf{B}$. Images are denoted with italic uppercase letters such as $\it{I}$. Specifically, the sign activation function is represented as $\mathbf{\it{sign}}$.
\subsection{Problem Definition}
Assume that $\mathbf{X} = \{\textit{I}_i\}^N_{i=1}$ is a training vehicle image set which contains $N$ training instances. $\mathbf{Q} = \{I_i^q\}^{N_{q}}_{i=1}$ and $\mathbf{G} = \{I^g_i\}^{N_{g}}_{i=1}$ are the query set and gallery set, respectively. $\mathbf{Y} = \{\textit{y}_i\}^N_{i=1} \in \{0,1\}^C$ is the corresponding label set and $C$ is the number of identities. The goal of deep hash learning for vehicle re-identification is to learn a non-linear hash function $\mathcal{H}: I \mapsto \mathbf{b} \in \{0,1\}^K $ which takes as input an image $\it{I}$ and map it into a K-bit binary hash code vector $\mathbf{f}$ such that the visual similarity in the raw pixel space is preserved for each vehicle image in the hamming space. The whole learning-based hash function could be formulated as:
\begin{equation*}
  \mathbf{b}  = \mathcal{H}(I) = \mathbf{\it{sign}} (\mathcal{F}_{hash}(\textit{I})) 
\end{equation*}
where $\mathcal{F}_{hash}$ stands for the non-linear deep neural network responsible for generating hash codes. $\textit{I}$ is an image. $\mathbf{\it{sign}(v)}$ returns 1 if $v \geq 0$ and returns -1 if $v < 0$. 
\section{Proposed Model}

\subsection{Hashing Layer}
We adopt \textbf{ResNet-50}~\cite{he2016deep} as the backbone feature extraction network, which is denoted as $\mathcal{F}_{\theta}$. The feature output vector $\mathbf{f_i}$ for an image $I_i$ from $\mathcal{F}_{\theta}$ is denoted as:
\begin{equation}
    \mathbf{f}_i = \mathcal{F}_{\theta}(I_i)
\end{equation}

To generate a binary hashing code of diversified lengths, we propose to devise a hashing layer on top of the backbone network. Concretely, following the global average pooling layer, we add a fully connected layer taking as input a feature vector $\mathbf{f}$ of $M$ dim and output K-dim continuous hash vector $\mathbf{h}$:
\begin{equation}
    \mathbf{h} = \mathcal{F}_{\theta_1}(\mathbf{f}) = \mathbf{f}W_1^T + \textbf{b}
\end{equation}
where $W_1 \in \mathbb{R}^{M\times K}$ and $\mathbf{b} \in \mathbb{R}^{K}$ denotes the weight and bias parameters while $B$ is the length of the \textit{Hamming} hashing code.
\subsection{Similarity-preserving Feature Learning}
We organize the input vehicle into triplet samples and formulate the hash problem as a similarity preserving problem. Concretely,~inspired by \cite{hermans2017defense},~we form input batches by first randomly sampling $P$ vehicle identities, then, we randomly sample $K_1$ instances for each vehicle. Thus, we get a batch of  $P*K_1$ images in total. For every image $I_i$ in the batch, we will use it as an anchor image and then find an image $I_i^+$ which is the hardest positive sample for the anchor in the batch. Analogously, we can select the hardest negative sample for the anchor $I_i$ in the batch, which is the image belonging to another vehicle but closest to the anchor $I_i$ in the batch according to a predefined distance metric $D$. In doing so, a training set $\mathbf{T_{1}} = \{(I_i, I_i^+, I_i^-)\}_{i=1}^{P*K_1}$ is derived. For each image $I$, the corresponding continuous hash vector $h$ is computed with the hash function $\mathcal{H}$:
\begin{equation}
    h_i = \mathcal{H}(I_i) = \mathcal{F}_{\theta_1}(\mathcal{F}_{\theta}(I_i))
\end{equation}
where $h_i$ is the continuous hash vector for image $I_i$ and $\mathcal{F}_{\theta}$ denotes the backbone non-linear convolutional neural network structure. $\mathcal{F}_{\theta_1}$ denote the fully connected feed-forward neural network (hash layer) for hash code generation.~Since the goal is to preserve the similarity of the images in the hamming space, it is natural to propose the following constraint:
\begin{equation}
    \mathcal{D}_H(h_i, h_i^+) < \mathcal{D}_H(h_i, h_i^-) 
\end{equation}
where $\mathcal{D}_H$ represents the hamming distance metric. The intuition is that the hamming distance between the anchor and the hardest positive sample should be smaller than the negative counterpart. We observe a linear relation between the \textit{Hamming} distance and the inner product of two vectors:
\begin{equation*}
    \mathcal{D}_H(h_i,h_j) = \frac{1}{2}(B - \langle h_i,h_j \rangle)
\end{equation*}
Thus, in this paper, we adopt inner product as a nice surrogate for \textit{Hamming} distance calculation.

In this way, we can finally formulate the triplet-based similarity preserving loss as:
\begin{equation}
\begin{array}{l}
\begin{split}
L_{Triplet} (h)
&= \sum_{i \in \mathbf{T_{1}}} [\alpha + \mathcal{D}_{H}(h_i,(h_i)^+ - \mathcal{D}_{H}(h_i,(h_i)^-]  \\
&=  \sum _{i=1}^{P} \sum _{a=1}^{K}[\alpha+\overbrace{\max _{p=1 \ldots K}\left\|e_{i}^{a}-e_{i}^{(p)}\right\|_{2}}^{\text {hardest positive}}  \\
&- 
\underbrace{\min _{n=1 \ldots K \atop j=1 \ldots P} \|e_{i}^{a}-e_{j}^{(n)}\|_{2} }_\textrm{hardest negative}]
\end{split}
\end{array}
\label{triplet}
\end{equation}
where $P$ denotes the number of vehicle identities in the batch and $K$ stands for the number of instances for each identity in the batch. $e_i^a$ denotes the anchor's vector belong to identity $i$. $\alpha$ is the margin parameter for the triplet loss. \par

On top of the triplet loss $L_{Triplet}$, to learn more discriminative features, we further add an \textbf{Identity Loss} on the feature output $\mathbf{f}$. To this end, another fully connected layer $\mathcal{F}_{\theta_2}$ of $N_C$ output logits is added after the $\mathbf{f}$ as shown in Fig.~\ref{fig:archi}. $N_C$ is the number of vehicle identities for the training images. By regarding each vehicle identity as a class, we can formulate the identity loss $L_{Identity}$ as:
\begin{equation}
L_{Identity}(f)=-\sum_{i=1}^{P*K} \log \frac{e^{{W}_{y_{i}}^{T} \mathbf{f}_i }}{\sum_{k_1=1}^{N_C} e^{{W}_{k_1}^{T} \mathbf{f}_{i}}}
\label{identity}
\end{equation}
where $W_k$ is the weight vector for identity $k$, $P * K_1$ is the number of images in the batch and $N_C$ is the number of identities in the batch.

\subsection{Discrete Hashing Learning}
The core assumption is that the learned binary \textit{Hamming} codes should preserve the label information. Specifically, for each image $I_i$, the binary \textit{Hamming} code $b_i$ is obtained by using the \textit{sign} activation function 
$
\operatorname{sgn}(x)=\left\{\begin{array}{c}
1, x>=0 \\
-1, x \le 0 
\end{array}\right.
$ on the continuous hash vector $h_i$ as:
\begin{equation}
    \mathbf{b}_i = \textit{sign}(h_i) = \textit{sign}(\mathcal{H}(\textit{I}_i))
\end{equation}

Similar to \cite{li2017deep}, we adopt a simple linear classifier to reconstruct the input ground-truth labels with the binary codes as:
\begin{equation}
    y_i = W^T_h\mathbf{b}_i
\end{equation}
where $y_i \in \{0,1\}^C$, $C$ is the number of identities, $W_h$ is the weight of the classifier for the binary codes $\mathbf{b}$. Then, we could formulate the discrete hashing loss as:
\begin{equation}
    L_{Quant}(\mathbf{b}) =  \mu\sum_{i=1}^{N}\left\|y_{i}-W^{T}_h \mathbf{b}_{i}\right\|_{2}^{2}+\nu \|W_h\|_{F}^{2}  
\end{equation}
where $\|\cdot\|_{2}$ is the $l2$ norm of a vector while $\|\cdot\|_{F}$ is the Frobenius norm of a specific matrix.

\subsection{Alternative Optimization}
We formulate the final total loss function as:
\begin{equation}
    L_{Total} = \lambda L_{Triplet}(\mathbf{h})  + \sigma L_{Identity}(\mathbf{f}) + L_{Quant}(\mathbf{b})
    \label{eq:total}
\end{equation}
where $\lambda$, $\beta$ and $\sigma$ are three predefined coefficients for controlling the influence of each individual loss. One might ponder the possibility of optimizing the Eq.~\ref{eq:total} with standardized stochastic gradient descent. Nonetheless, the \textit{sign} function is non-differentiable which can not be directly optimized by  back-propagating the gradients. Thus, the third term $L_{Quant}(\mathbf{b})$ could not be optimized in a same way as the first two terms. We, thus, unfold the last term in the Eq.~\ref{eq:total}:
\begin{equation}
 \begin{aligned}
  L_{Total} &= \lambda L_{Triplet}(\mathbf{h})  + \sigma L_{Identity}(\mathbf{f})) \\
  &+\mu \sum_{i=1}^{N}\left\|y_{i}-W_h^{T} b_{i}\right\|_{2}^{2}+\nu\|W_h\|_{F}^{2} 
 \end{aligned}
    \label{eq:totalfold}
\end{equation}

To circumvent the vanishing gradient problem introduced by \textit{sign}, some previous works adopts \textit{tanh} or \textit{sigmoid} activation function to generate continuous outputs which are the relaxation of the binary \textit{Hamming} codes in the training stage. In the testing stage, \textit{sign} activation function is adopted to obtain the corresponding binary \textit{Hamming} hashing codes.  However, such an optimization scheme is optimizing a relaxed surrogate of the original target and will result in learning sub-optimal \textit{Hamming} hashing codes owing to the sizable \textbf{Quantization Error}~\cite{liu2016deep}. \par
In this regard, we propose to adopt a novel \textbf{Alternative Optimization} scheme which considers the discrete nature of binary codes by constraining the outputs from the deep neural network to be binary codes directly. Similar to \cite{wang2016deep,li2017deep},~we introduce an auxiliary variable and derive an approximation of Eq.~\ref{eq:totalfold} as:
\begin{equation}
    \begin{aligned}
L_{Total} &= L_{Triplet}(h) + L_{Identity}(f) \\
&+\mu \sum_{i=1}^{N}\left\|y_{i}-W_h^{T} b_{i}\right\|_{2}^{2}+\nu\|W_h\|_{F}^{2} \\
&\text { s.t. } \quad \mathbf{b}_{i}=\operatorname{sgn}\left(h_{i}\right), \quad \mathbf{h}_{i} \in \mathbb{R}^{K \times 1}, \quad(i=1, \ldots, N)
\end{aligned}
\label{eq:var}
\end{equation}
where $\mathbf{h} = \mathcal{F}_{hash}(f) $ is the continuous hashing vector, $\mathbf{f}$ is the feature output from the deep convolutional neural network and  $\mathbf{b}$ is the binarized \textit{Hamming} hashing code. The above equation is a constrained optimization problem, which could be solved with the \textbf{Lagrange Multiplier Method} as:
\begin{equation}
        \begin{aligned}
L_{Total} &= L_{Triplet}(h) + L_{Identity}(f) \\
&+\mu \sum_{i=1}^{N}\left\|y_{i}-W^{T}_h b_{i}\right\|_{2}^{2}+\nu\|W_h\|_{F}^{2} \\ 
         &+ \eta \sum_{i=1}^{N}\left\|\mathbf{b}_{i}-\operatorname{sgn}\left(\mathbf{h}_{i}\right)\right\|_{2}^{2} \\
&\text { s.t. } \quad \mathbf{b}_{i} \in \{-1,1\}^K, \quad \mathbf{h}_{i} \in \mathbb{R}^{K \times 1}, \quad(i=1, \ldots, N)
\end{aligned}
\label{eq:lagrange}
\end{equation}
The last term can also be viewed as a constraint which measures the quantization error between the continuous hashing feature vector and its binarized \textit{Hamming} code. Inspired by \cite{li2017deep}, we decouple the above optimization target Eq.~\ref{eq:lagrange} into two sub optimization problem, and solve them in an alternative fashion.\par
First, we fix $b$ and $W$, the optimization target is converted to 
\begin{equation}
\begin{aligned}
    L &= \lambda L_{Triplet}(\mathbf{h}) + \sigma L_{Identity}(\mathbf{f}) \\
    &+ \eta \sum_{i=1}^{N}\left\|\mathbf{b}_{i}-\mathbf{h}_{i}\right\|_{2}^{2}
\end{aligned}
\label{eq:sgd}
\end{equation} \par
Then, the parameters of the backbone $\theta$, hash layer $\theta_1$, and the identity loss $\theta_2 $ could be updated with standard \textbf{Stochastic Gradient Descent} and \textbf{Back Propagation} algorithm. \par
Secondly, we fix the parameters $\theta$, $\theta_1$, $\theta_2$ and $\mathbf{b}_i$ and the problem is converted into:
\begin{equation}
    L = \mu \sum_{i=1}^{N}\left\|y_{i}-W_h^{T} b_{i}\right\|_{2}^{2}+\nu\|W_h\|_{F}^{2} \\ 
    \label{eq:least}
\end{equation}
It is obvious to note that this is a \textbf{Least Squares problem} and has a closed-form solution as:
\begin{equation}
    W_h=\left(B B^{T}+\frac{\nu}{\mu} I\right)^{-1} B^{T} Y
    \label{eq:least}
\end{equation}
where $B = \{\mathbf{b}_i\}_{i=1}^N \in\{-1,1\}^{K \times N}$ and $Y = \{y_i\}^N_{i=1} \in \mathbb{R}^{C \times N}$.
\par
Finally, after solving $W_h$, when we fix the parameters $\theta$, $\theta_1$, $\theta_2$ and $W_h$, we obtain the final objective as:
\begin{equation}
    \begin{aligned}
&L=\mu \sum_{i=1}^{N}\left\|y_{i}-W_h^{T} \mathbf{b}_{i}\right\|_{2}^{2}+\eta \sum_{i=1}^{N}\left\|\mathbf{b}_{i}-\mathbf{h}_{i}\right\|_{2}^{2} \\
&\text { s.t. } \quad \mathbf{b}_{i} \in\{-1,1\}^{K},(i=1, \ldots, N)
\end{aligned}
\end{equation}
Inspired by \cite{li2017deep}, we adopt the discrete cyclic coordinate descent method to solve $B$ row by row in an iterative fashion as:
\begin{equation}
    \min _{B}\left\|W_h^{T} B\right\|^{2}-2 \operatorname{Tr}(P), \quad \text { s.t. } B \in\{-1,1\}^{K \times N}
    \label{eq:b}
\end{equation}
where $P=W_h Y+\frac{\eta}{\mu} H$. The overall training algorithm for \textbf{DVHN} is illustrated in Algo.~\ref{ag:first}.

\begin{algorithm}
 \caption{Optimization Algorithm for DVHN}
 \label{ag:first}
  \KwIn{Training vehicle image set $\mathbf{T}$, labels $\mathbf{Y}$, ~learning rate $\epsilon$ and parameters $\lambda$, $\beta$, $\sigma$,~ the margin parameter~$\alpha$, ~maximum training iteration $T$}
  \KwOut{network parameters $\theta$, $\theta_1$, $\theta_2$, $W_h$, $\mathbf{B}$;}
   \textbf{Initialize:} the resnet-50 network parameters $\theta$ with pretrained weights on ImageNet, $\theta_1, \theta_2, W_h, B$ with random values, and t = 0\\
  \While{not converged and $t < T$ }
  {
    $t = t +1 $\;
    \For{$i\gets0$ \KwTo $100$ }{
    Freeze $W, b$ \;
    Sample $P$ vehicle identities. Then sample $K$ instances for each $p\in P $, resulting in a batch image set $\mathbf{B} = \{I_i\}_{i=1}^{P*K}$ and the corresponding label set $\mathbf{Y} = \{y_i\}_{i=1}^{p*K}$\;
    For each image $I_i$, calculate $f_i = \mathcal{F}_\theta(I_i)$                                 \;
    Calculate continuous hash vector $h_i = \mathcal{F}_{\theta_1}(f_i)$  \;
    Obtain loss $L_{Identity}$ by Eq. \ref{identity}   \;
    Construct triplet set $\{h_i, h_i^+, h_i^-\}$   \;
    Obtain loss $L_{Triplet}$ by Eq.~\ref{triplet}  \;
    
    Use \textbf{SGD} to get gradients $\nabla_{\theta_*}$~$*\in \{\theta,\theta_1,\theta_2\}$ for Eq.~\ref{eq:sgd}\;
    Update $ \theta_* = \alpha ( \theta_* - \nabla_{\theta_*}) $ \;
    }
    Freeze $\theta,~\theta_1,~\theta_2$, update $W_h$ with Eq.~\ref{eq:least} \;
    Freeze $\theta,~\theta_1,~\theta_2,W_h$, update $\mathbf{B}$ through Eq.\ref{eq:b} 
  }
\end{algorithm}

\subsection{Retrieval Process}
In this section, we elaborate on how to perform efficient vehicle re-identification given the trained model.
Generally, we are given a query image set $\mathbf{Q}$ and a gallery image set $\mathbf{G}$. For each query image $q_k$, the task is to identify the images in $\mathbf{G}$ belonging to the same vehicle. In the testing phase, for each query image $q_k$, we obtain the discrete binary code vector $\mathbf{b}_k^q$ through:
\begin{equation}
    \mathbf{b}_k^q = \mathbf{\it{sign}(\mathcal{F}_{\theta_1}(\mathcal{F}_{\theta}(\it{I}^q_k)))}
\end{equation}
Subsequently, in a same fashion, for all the images in $\mathbf{G} = \{\it{I}^g_k\}_{k=1}^{N_g} $, we obtain a set of binary code vectors $\mathbf{H^g} = \{\mathbf{b}_k^g\}_{k=1}^{N_g} $. Then, we can rank the gallery code vectors through their hamming distance with respect to the query image code vector. We note that there is a linear relationship between inner product $\langle\cdot, \cdot \rangle$ and hamming distance $\mathcal{D}_H$, which is stated as:
\begin{equation*}
    \mathcal{D}_H(\mathbf{b}_i,\mathbf{b}_j) = \frac{1}{2}(B - \langle \mathbf{b}_i, \mathbf{b}_j \rangle)
\end{equation*}
where $K$ is the bit length of the binary hashing code $\mathbf{b}$. The above equation makes the inner product a nice surrogate for hamming distance computing. In the later stage, we compute the hamming distance between $\mathbf{b}_k^q$ and all the gallery hash codes in $G$. Then, we sort the distances in an increasing order. Finally, we rank the images in $\mathbf{G}$ according to the sorted hamming distances.

\section{Experiments}
\subsection{Implementation Details}
We implement our model with pytorch~\cite{paszke2019pytorch}. The detailed framework is illustrated in Fig.~\ref{fig:archi}. Specifically, after the final $f_{avg}$ layer of resnet50, we add a fully connected layer of $K$ output logits where $K$ denotes the hash code length. For the classification stream, we added another fully connected layer of $N_C$ output logits, $N_C$ denoting the number of classes. These two additional layers are initialized with "normal" initialization, the mean being $0$ and standard deviation being $0.01$. In the training process, the number of identities $P$ is set to 16, and the number of instances is $6$, resulting in a batch of 96 images. During training, the parameters are optimized with "amsgrad" optimizer. The initial learning rate $\epsilon$ is set to $3e-4$, the weight decay to $\it{5e-4}$, the beta1 to 0.9, and the beta2 to 0.99. The margin parameter $\alpha$ in Eq.~\ref{triplet} is set to $0.3$. For the coefficients controlling the influence of each loss $(\lambda,\beta,\sigma)$, we empirically set all of them to $1$

\subsection{Datasets and Evaluation Protocols}
\textbf{Datasets.} We conduct experiments on two widely-studied public datasets: \textbf{VehicleID}~\cite{liu2016deeprela} and \textbf{VeRi}~\cite{liu2016deep}. \textbf{VehicleID} is a large-scale vehicle re-identification dataset containing 221763 images of 26267 vehicle identities in total. Among all the images, 110,178 images of 13,134 vehicles are selected for training. For testing,~three subsets which contain 800, 1600, 2400 vehicles are extracted from the testing dataset and are denoted as \textit{800}, \textit{1600}, \textit{2400}, respectively. We empirically evaluated the performances of our model on \textit{800} and \textit{1600} scenario. Specifically, for \textit{800}, the query set contains 5693 images of 800 vehicle identities, and the gallery dataset contains 800 images in total, one for each vehicle identity. In a similar fashion, for the \textit{1600} scenario, the query set comprises 11777 images of 1600 cars, while the gallery set contains 1600 images of 1600 identities.  \textbf{VeRi} is a real-world urban surveillance vehicle dataset containing over 50000 images of 776 vehicle identities, among which 37781 images of 576 vehicles are designated as the training set. For testing, 1678 images of 200 vehicles are designated as the query set, while the gallery contains 11579 for the same 200 vehicles. \\
\textbf{Evaluation Protocols.} We evaluate our model on two widely-adopted metrics: cumulative matching characteristics (CMC) and mean average precision (mAP). The CMC score computes the probability that a query image appears in a different-sized retrieval list. However, it only counts the first match, making it unsuitable in a scenario where the gallery contains more than one image for each query. Thus, we also employ \textbf{mAP}, which, essentially, takes into consideration both the \textit{precision} and \textit{recall} of the retrieved results.

\begin{table*}
\newcolumntype{Y}{>{\centering\arraybackslash}X}
\newlength\mylength
\setlength\mylength{\dimexpr 0.8\textwidth-2\tabcolsep}

    \caption{Vehicle Re-identification Results of State-of-The-Art Deep Hash Methods on \textbf{Vehicle (800} and \textbf{VeRi}}
    
    \begin{tabularx}{\textwidth}{ XX| Y|Y|Y|Y || Y|Y|Y|Y } \hline
    \multicolumn{2}{l|}{Datasets} & \multicolumn{4}{c||}{VehicleID (\textit{800})} &\multicolumn{4}{c}{VeRi } 
       \\ \hline
    \multicolumn{2}{l|}{Methods} & 256 & 512 & 1024 & 2048  & 256 &512 &1024 & 2048 \\\hline
    
    \multicolumn{2}{l|}{SH\cite{weiss2008spectral} (NeurIPS)} & 23.98 & 23.64 & 21.09 & 17.71 &3.33 & 2.75 & 2.08 & 1.57 \\
    \multicolumn{2}{l|}{ITQ\cite{gong2012iterative} (TPAMI)} &  23.84 & 25.39 & 25.93 & 26.26 & 4.80 & 4.95 & 5.04 & 5.14 \\
    \hline
    \hline
    \multicolumn{2}{l|}{DSH\cite{liu2016deep12} (CVPR)} & 30.56 &  32.53 & 34.22 & 33.62 & 8.93 & 9.77 & 10.24 & 10.47 \\
  
    \multicolumn{2}{l|}{DHN\cite{zhu2016deep} (AAAI)} & 43.57 & 45.42 & 46.70 & 46.39 & 13.85 & 13.93 & 14.31 & 13.71\\ 
  
    \multicolumn{2}{l|}{HashNet\cite{cao2017hashnet} (ICCV) }& \textcolor{blue}{62.95} & \textcolor{blue}{64.52} & \textcolor{blue}{65.96} & \textcolor{blue}{66.61} & \textcolor{blue}{27.13} & \textcolor{blue}{28.40} & \textcolor{blue}{29.94} & \textcolor{blue}{29.30} \\
    \multicolumn{2}{l|}{DCH\cite{cao2018deep} (CVPR) }& 37.08 & 34.55 & 26.04 & 24.13 & 19.75 & 19.23 & 12.15 & 14.96 \\
    \multicolumn{2}{l|}{DPN\cite{fan20deep} (IJCAI)}& 57.16 & 61.92 & 63.55 & 64.65 & 10.03 & 11.26 & 13.52 & 13.58 \\
   
     \multicolumn{2}{l|}{DVHN  }&\textcolor{red}{\textbf{71.61}} & \textcolor{red}{\textbf{73.69}} & \textcolor{red}{\textbf{75.86}} & \textcolor{red}{\textbf{76.82}} & \textcolor{red}{\textbf{54.61}} & \textcolor{red}{\textbf{58.38}} & \textcolor{red}{\textbf{59.27}} & \textcolor{red}{\textbf{62.02}} \\
     \hline
     \hline
    \end{tabularx}
    \label{table:mainres}
\end{table*}

\begin{table}
\newcolumntype{Y}{>{\centering\arraybackslash}X}
\newlength\mylengthh
\setlength\mylengthh{\dimexpr 0.5\textwidth-2\tabcolsep}

    \caption{Vehicle Re-identification Results of State-of-The-Art Deep Hash Methods (\textbf{VehicleID (1600)})}
    
    \begin{tabularx}{0.5\textwidth}{ XX| Y|Y|Y|Y } \hline
    \multicolumn{2}{l|}{Datasets} & \multicolumn{4}{c}{VehicleID (\textit{800})} 
       \\ \hline
    \multicolumn{2}{l|}{Methods} & 256 & 512 & 1024 & 2048  \\\hline
    
    \multicolumn{2}{l|}{SH\cite{weiss2008spectral} (NeurIPS)} & 19.96 & 20.06 & 16.32 &13.83 \\
    \multicolumn{2}{l|}{ITQ\cite{gong2012iterative} (TPAMI)} &  20.23 & 21.45 & 21.27 & 22.23  \\
    \hline
    \hline
    \multicolumn{2}{l|}{DSH\cite{liu2016deep12} (CVPR)} & 26.25 &  29.39 & 30.36 & 30.97  \\
  
    \multicolumn{2}{l|}{DHN\cite{zhu2016deep} (AAAI)} & 39.97 & 41.77 & 43.34 & 43.72 \\ 
  
    \multicolumn{2}{l|}{HashNet\cite{cao2017hashnet} (ICCV) }& \textcolor{blue}{56.89} & \textcolor{blue}{69.99} & \textcolor{blue}{60.05} & \textcolor{blue}{60.88} \\
    \multicolumn{2}{l|}{DCH\cite{cao2018deep} (CVPR) }& 27.23 & 25.33 & 19.67 & 18.45  \\
    \multicolumn{2}{l|}{DPN\cite{fan20deep} (IJCAI)}& 53.68 & 58.03 & 60.23 & 61.42  \\
   
     \multicolumn{2}{l|}{DVHN  }&\textcolor{red}{\textbf{66.67}} & \textcolor{red}{\textbf{69.12}} & \textcolor{red}{\textbf{71.83}} & \textcolor{red}{\textbf{72.72}}  \\
     \hline
     \hline
    \end{tabularx}
    \label{table:mainres1}
\end{table}

\subsection{Comparisons with State-of-the-arts}
In this section, we compare the results of our proposed \textbf{DVHN} and the state-of-the-art deep hash methods. Specifically, the competing methods could be divided into two categories: unsupervised hash methods and supervised hash methods. For the unsupervised methods, we include \textbf{SH}~\cite{weiss2008spectral} and \textbf{ITQ}~\cite{gong2012iterative}. Note that these two methods are reliant upon extra feature extractors to get feature representations for the dataset. To this end, we adopt \textit{AlexNet} which is pretrained on \textbf{ImageNet} to extract features. For the supervised setting, we further incorporate \textbf{DSH}~\cite{liu2016deep12} which was among the first works attempting to learn discrete hamming hash codes with deep convolutional neural networks. In addition, we incorporate other recent methods for detailed comparisons and analyses including ~\textbf{DHN}~\cite{zhu2016deep},~\textbf{HashNet}~\cite{cao2017hashnet},~\textbf{DCH},~\cite{cao2018deep}~\textbf{DPN}~\cite{fan20deep}. Note that we adopt original source codes for \textbf{SH}, \textbf{ITQ},~ \textbf{DCH} and \textbf{DHN}. For other methods, we implement them with \textit{pytorch}. We first provide analysis for the results of all the models with 2048 hash bits. It is rather evident from Tab.~\ref{table:mainres} and Tab.~\ref{table:mainres1} that traditional unsupervised methods like \textbf{SH} and \textbf{ITQ} exhibit significant poorer performance across two datasets. Their mAP scores on \textbf{VeRi} are 1.57\% and 5.14\% when the hashing bit length equals $2048$, respectively. On \textbf{VehicleID (800)} and \textbf{ VehicleID (1600)}, the performance are slightly better. Specifically, \textbf{Sh} achieves mAPs of 17.71\% and 13.83\% while \textbf{ITQ} reaches 26.26\% and 22.23\% for 2048 bits on these two datasets.~The unsatisfactory results could in large part be attributed to the fact that unsupervised methods could not learn the discriminative features, leading to sub-optimal hash codes. Clearly, supervised methods achieve better results across two datasets for all the hashing bit lengths. Especially, \textbf{HashNet} and \textbf{DPN} achieve pronounced results on \textbf{VehicleID (\textit{800})}, 66.61\% and 64.65\% for $2048$ bits, respectively. Across other hashing bit lengths, they are also clear winners regarding \textbf{mAP}. In terms of \textbf{VeRi}, poorer performance could be spotted, 29.30\% mAP for \textbf{HashNet} and 13.58\% for \textbf{DPN} on 2048 bits. The degraded performance could result from the small discrepancy of images from the same vehicle. Unsurprisingly, our proposed \textbf{DVHN} is a clear winner against all the competing methods across all the compared bit lengths. Concretely, we achieve the state-of-the-art results with pronounced margins of 11.05\% and 35.46\% in terms of \textbf{Rank@1} on \textbf{VehicleID} and \textbf{VeRi} on 2048 bits,~respectively. For 256 bits, we obtain 11.54\% and 30.93\% performance improvements of \textbf{Rank@1}. In terms of \textbf{mAP}, for 2048 bits, we achieve 76.82\%, 72.72\% and 56.57\% on \textbf{VehicleID (800)}, \textbf{VehicleID (1600)} and \textbf{VeRi}, outperforming the second-best results by 10.21\%, 11.30\% and 32.72\%. The superior performances collectively and evidently demonstrated the advantages of the joint hash and classification stream design, which enables the model to learn more discriminative features. \par

\begin{figure*}
\centering
\subfigure[CMC Results of  \textbf{VehicleID (800)} of \textbf{256} bits] 
{\includegraphics[width=5.8cm]{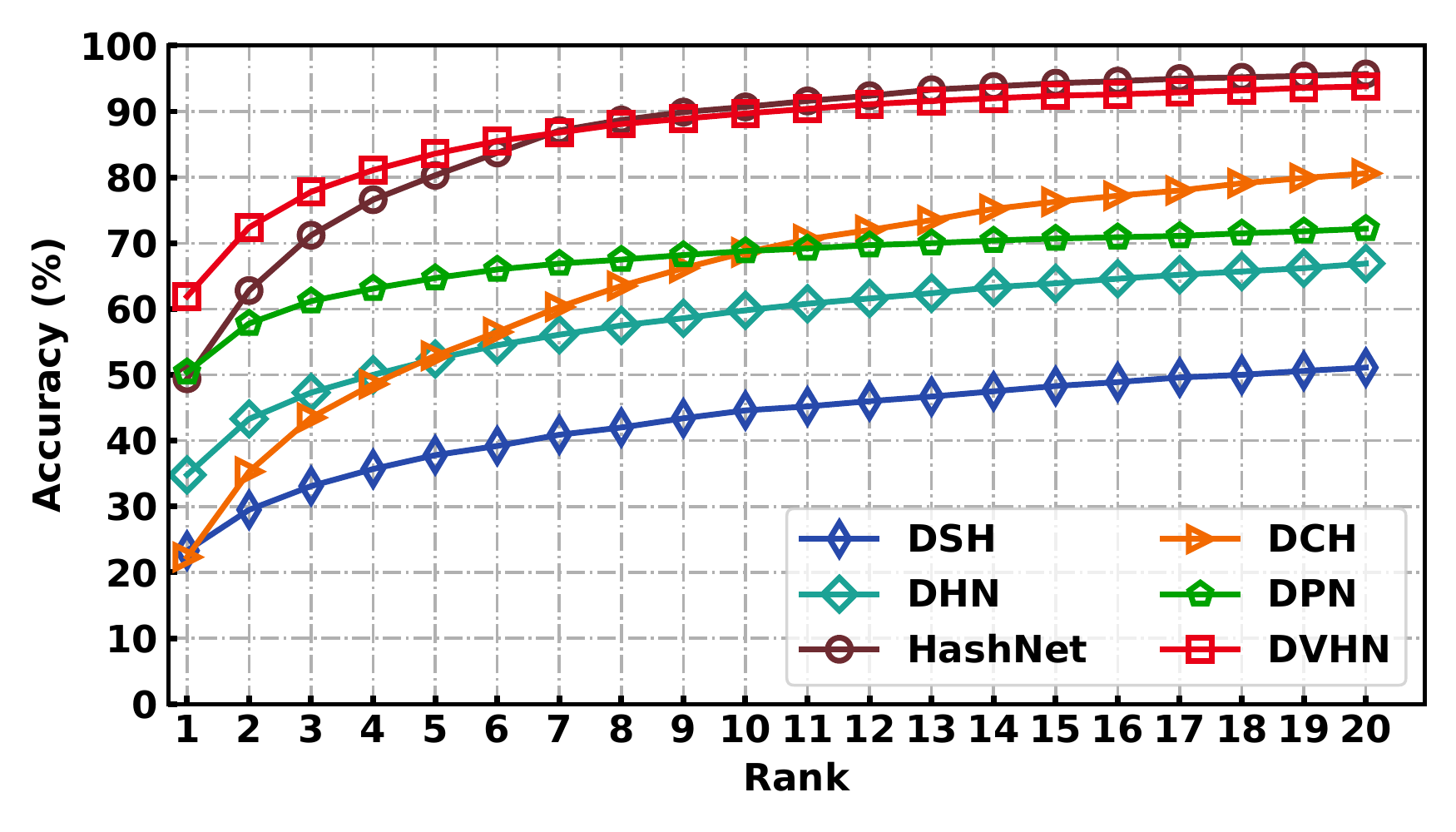}
}
\subfigure[CMC Results of  \textbf{VehicleID (800)} of \textbf{512} bits]{\includegraphics[width=5.8cm]{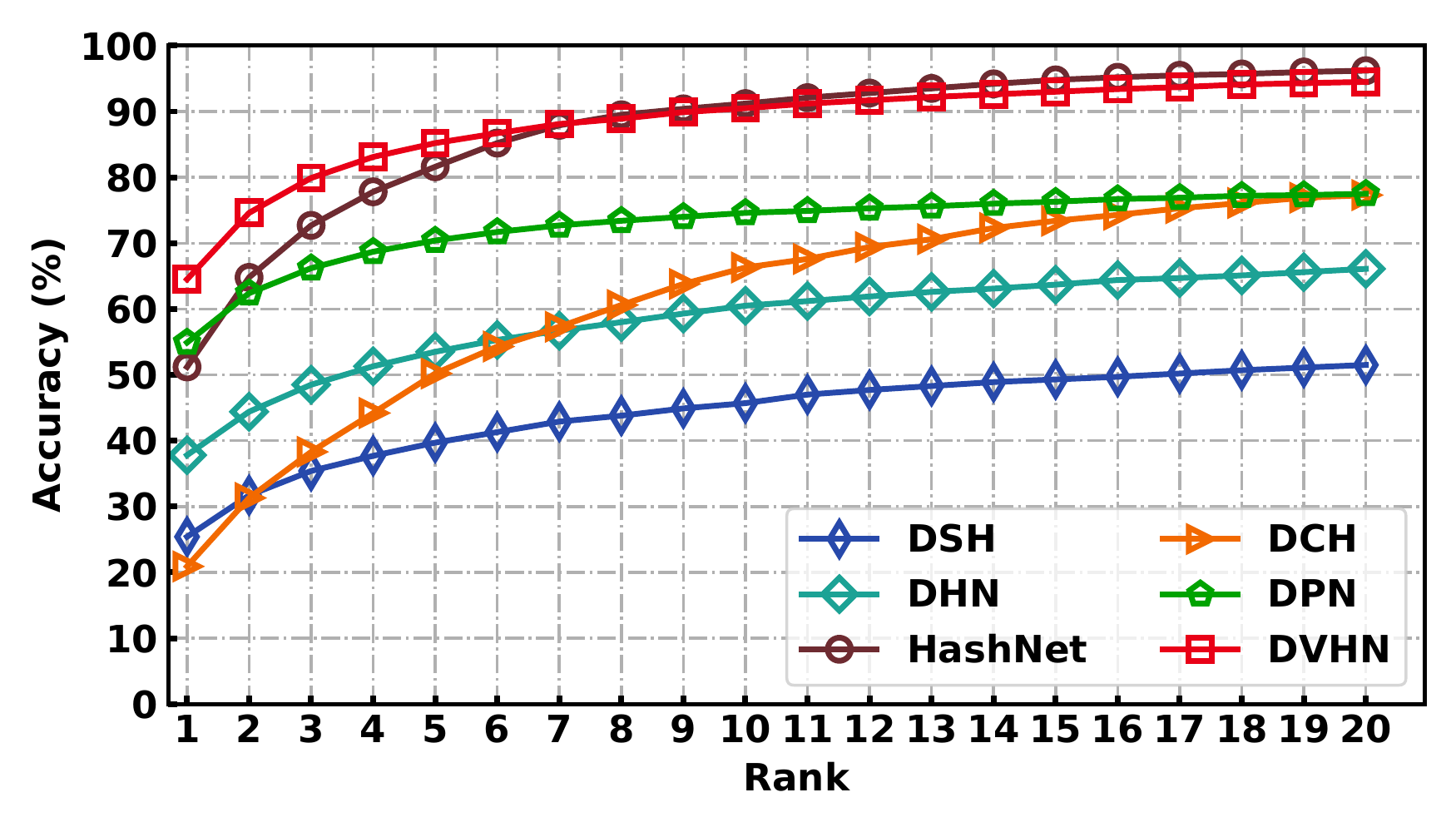}}
\subfigure[CMC Results of  \textbf{VehicleID (800)} of \textbf{1024} bits]{\includegraphics[width=5.8cm]{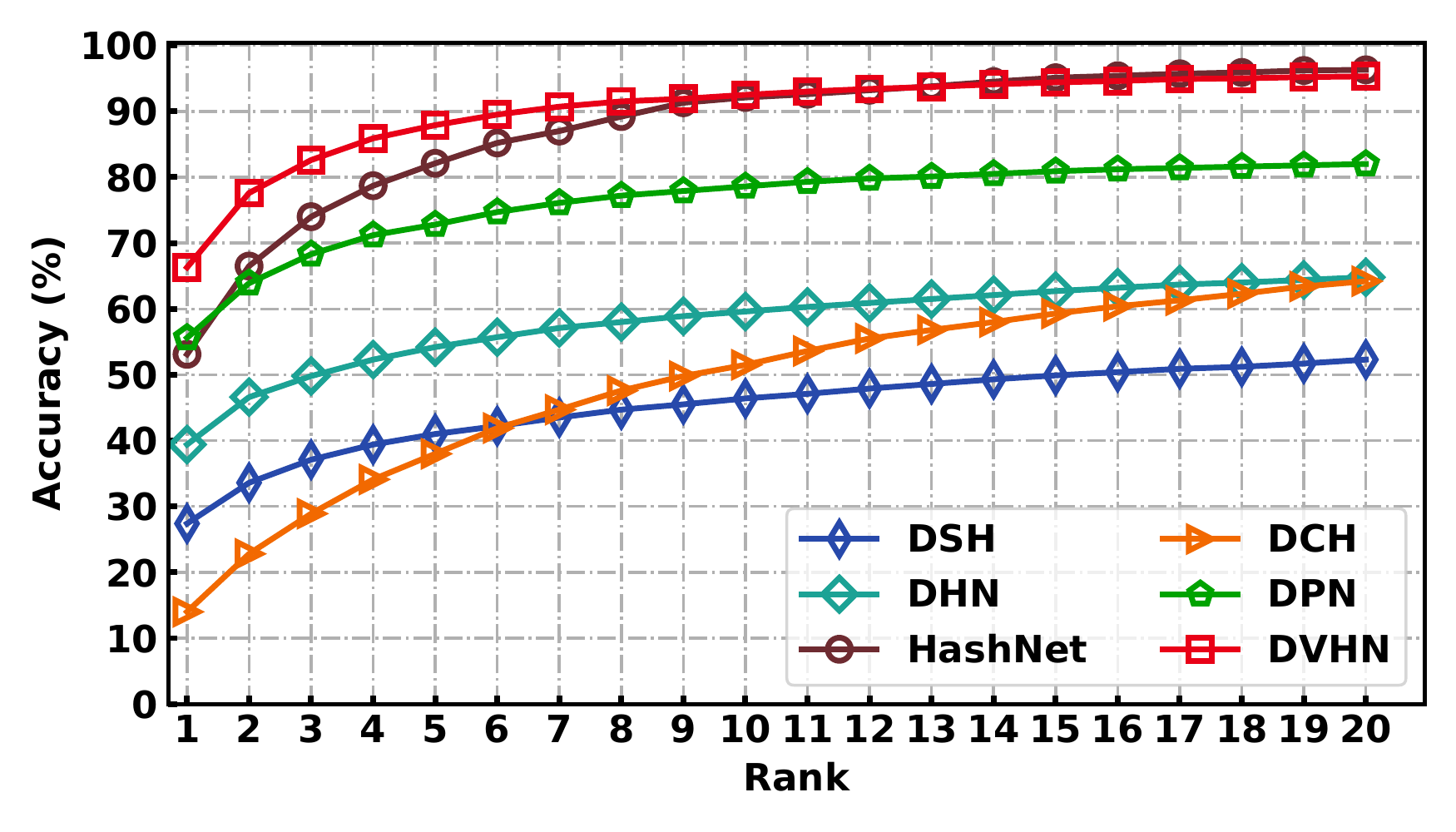}}
\\
\subfigure[CMC Results of \textbf{VehicleID (800)} of \textbf{2048} bits] 
{\includegraphics[width=5.8cm]{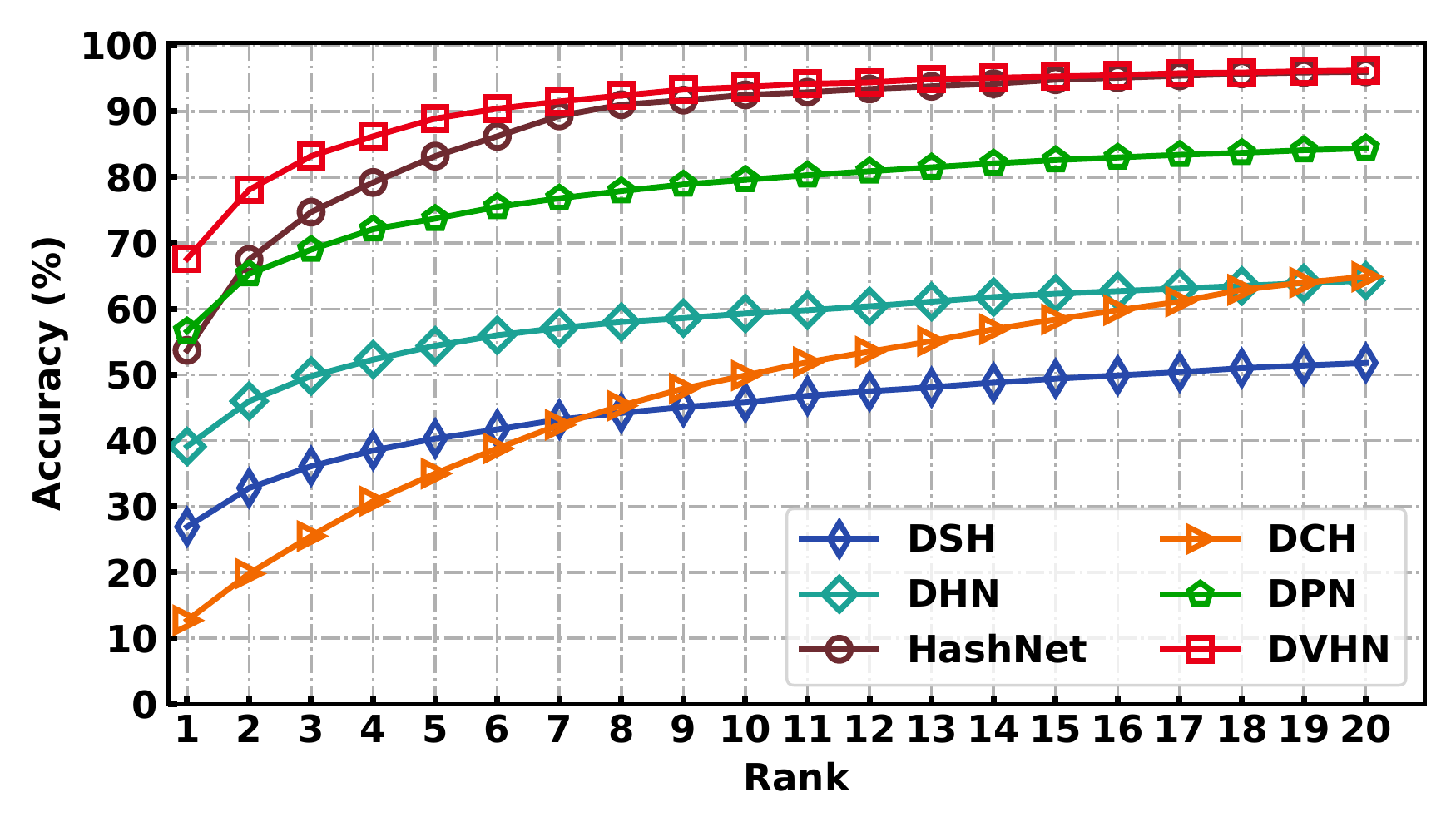}} 
\subfigure[CMC Results of  \textbf{VehicleID (1600)} of \textbf{256} bits]{\includegraphics[width=5.8cm]{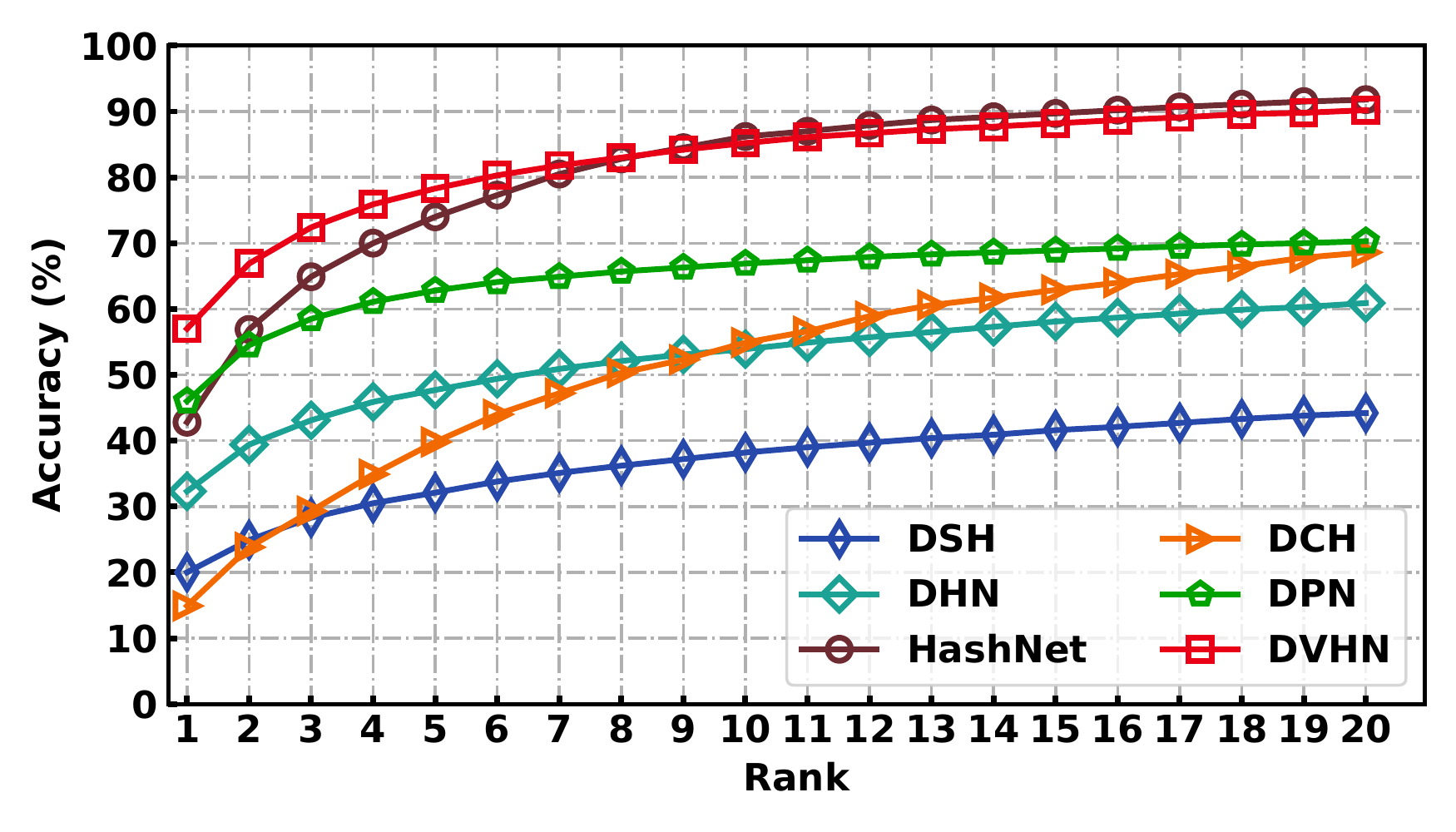}}
\subfigure[CMC Results of  \textbf{VehicleID (1600)} of \textbf{512} bits]{\includegraphics[width=5.8cm]{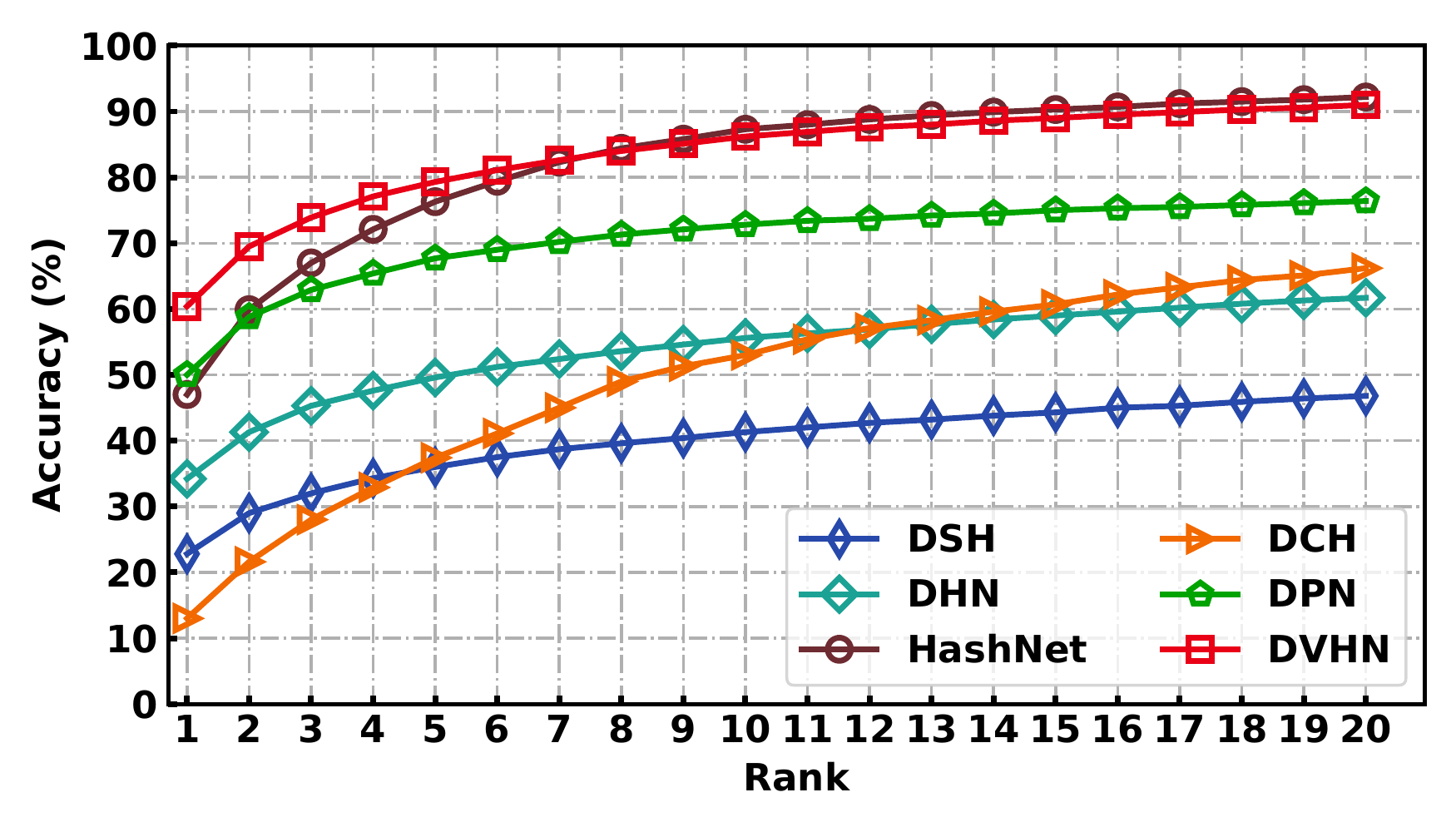}}
\\
\subfigure[CMC Results of  \textbf{VehicleID (1600)} of \textbf{1024} bits] 
{\includegraphics[width=5.8cm]{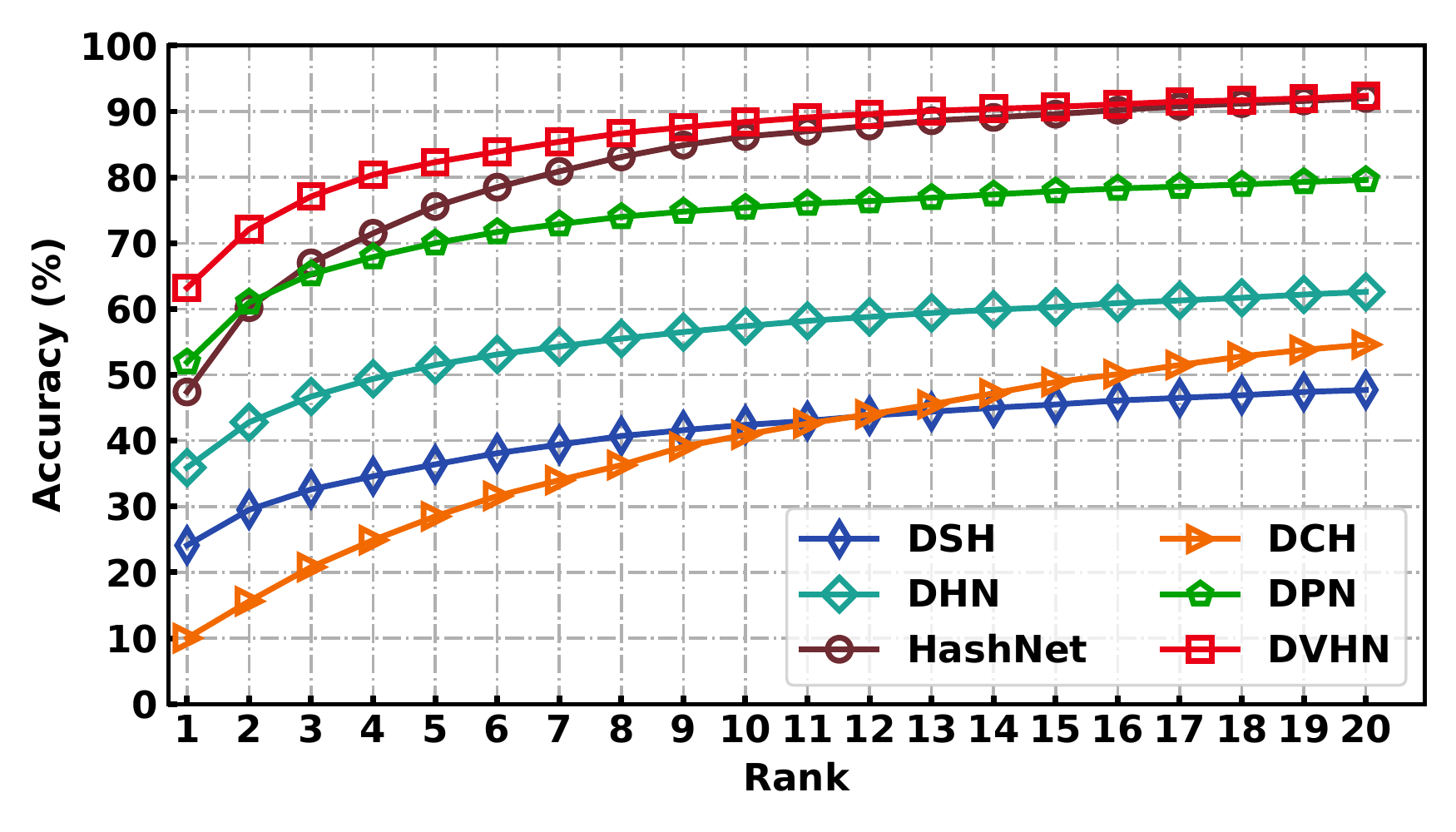}} 
\subfigure[CMC Results of \textbf{VehicleID (1600)} of \textbf{2048} bits]{\includegraphics[width=5.8cm]{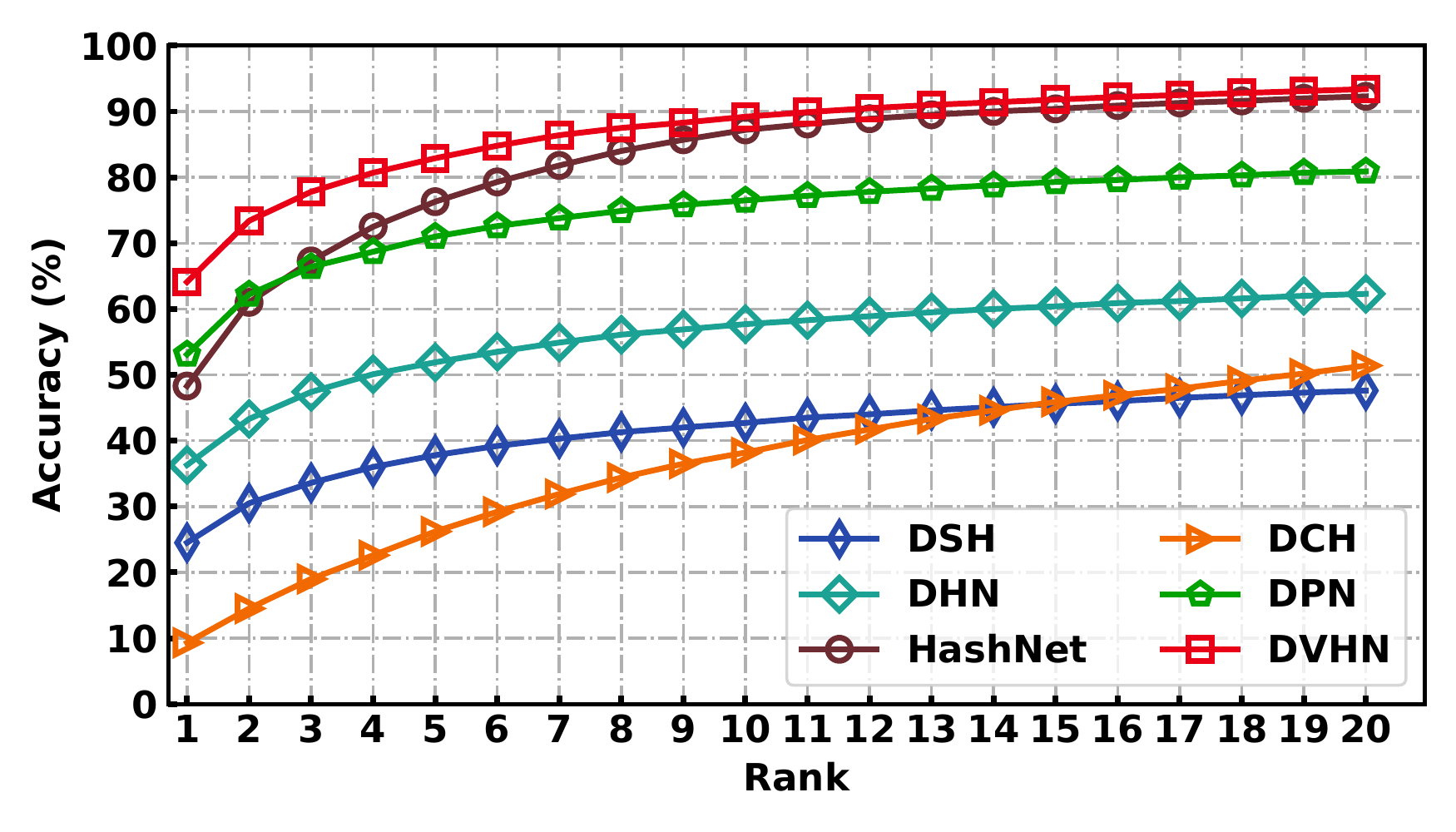}}
\subfigure[CMC Results of  \textbf{VeRi} of \textbf{256} bits]{\includegraphics[width=5.8cm]{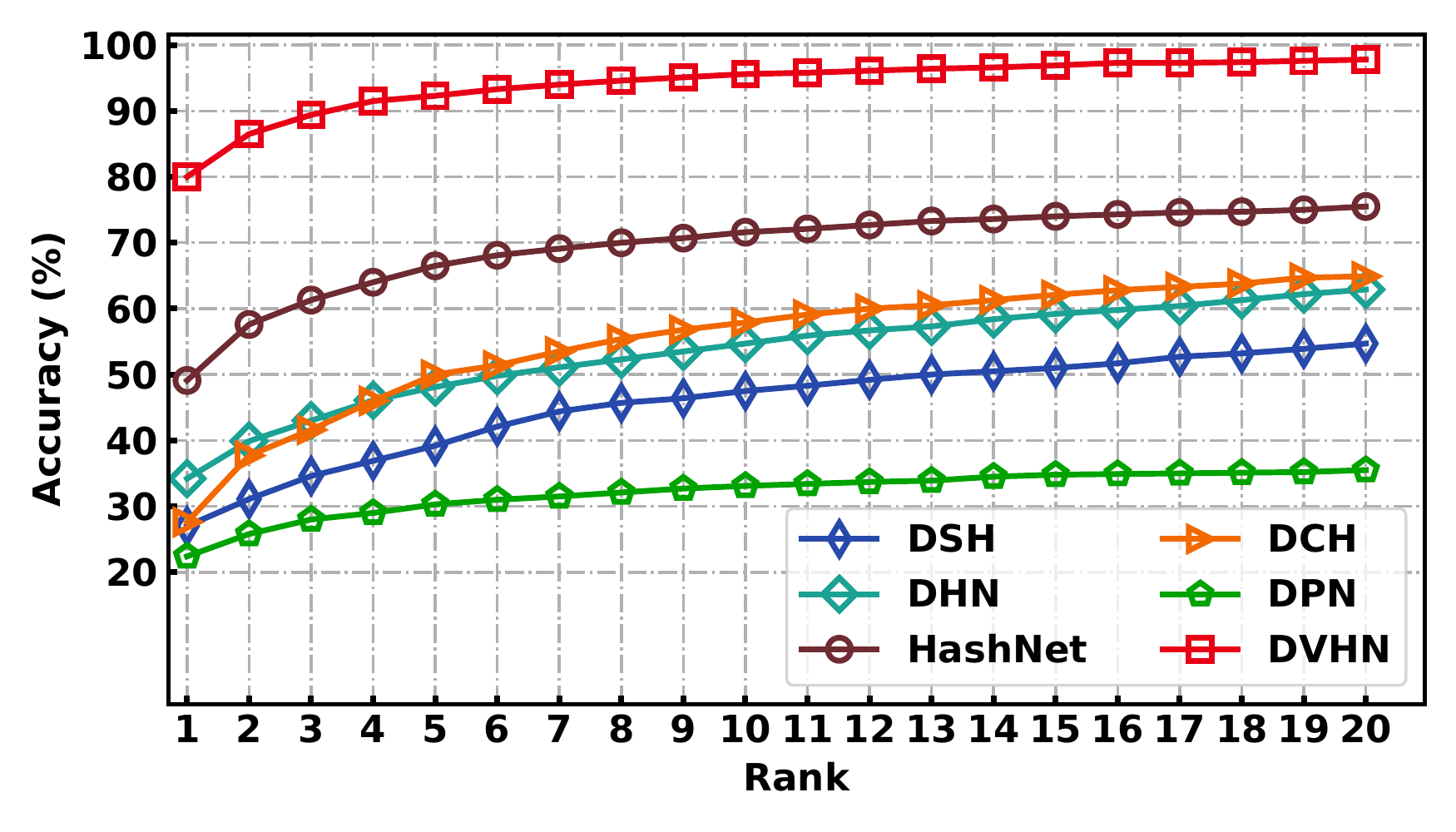}}
\\
\subfigure[CMC Results of  \textbf{VeRi} of \textbf{512} bits] 
{\includegraphics[width=5.8cm]{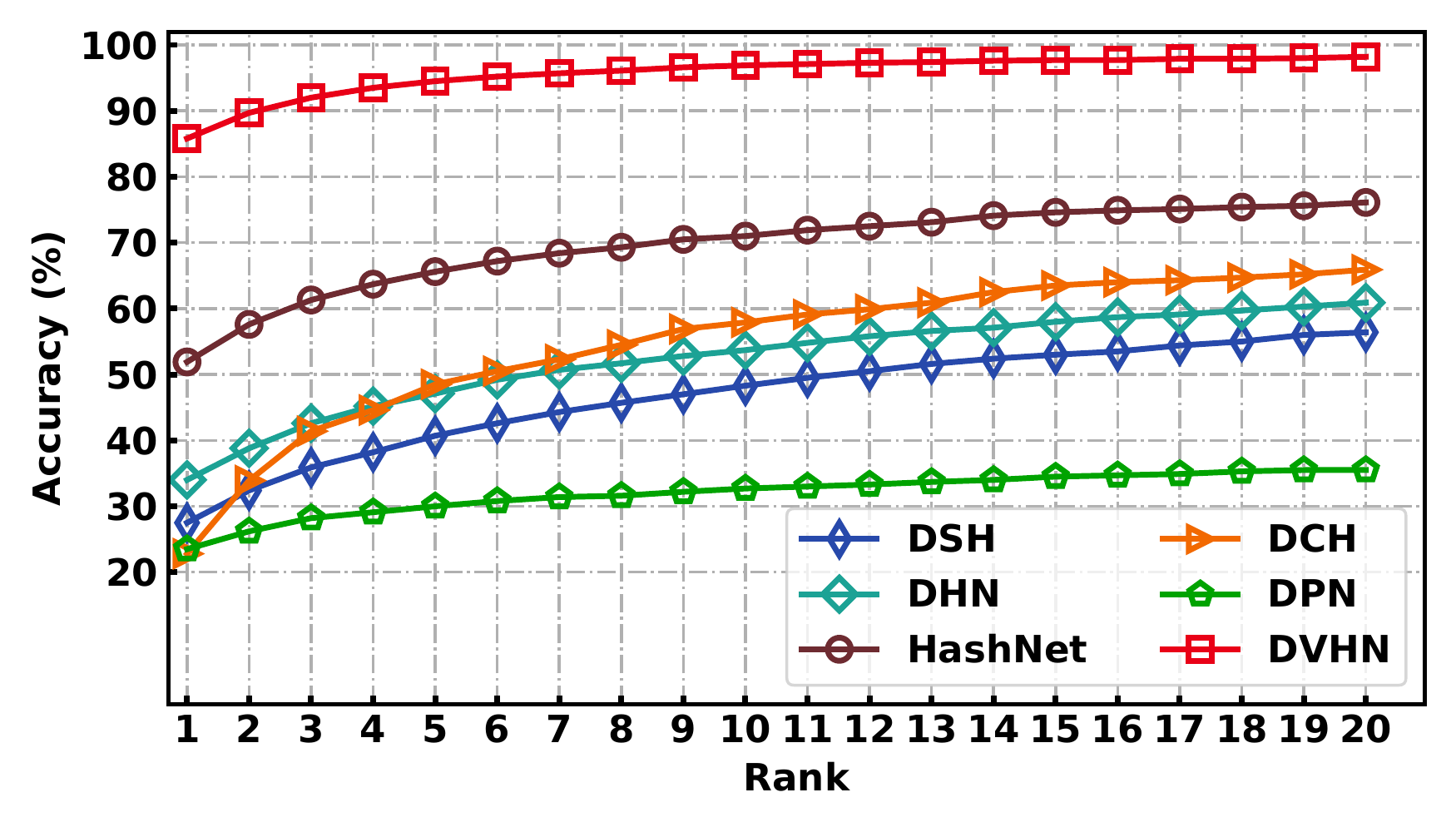}} 
\subfigure[CMC Results of  \textbf{VeRi} of \textbf{1024} bits]{\includegraphics[width=5.8cm]{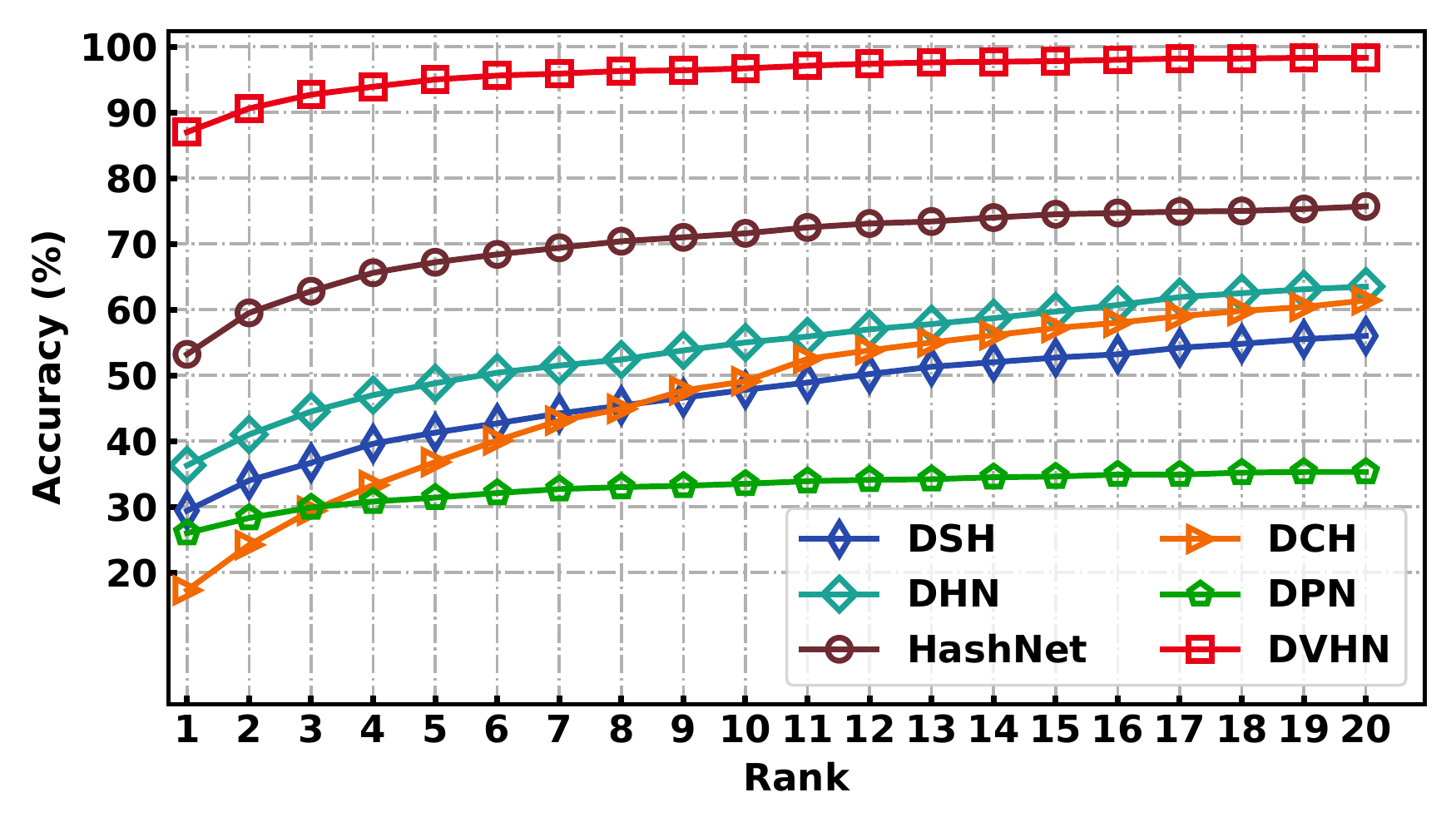}}
\subfigure[CMC Results of  \textbf{VeRi} of \textbf{2048} bits]{\includegraphics[width=5.8cm]{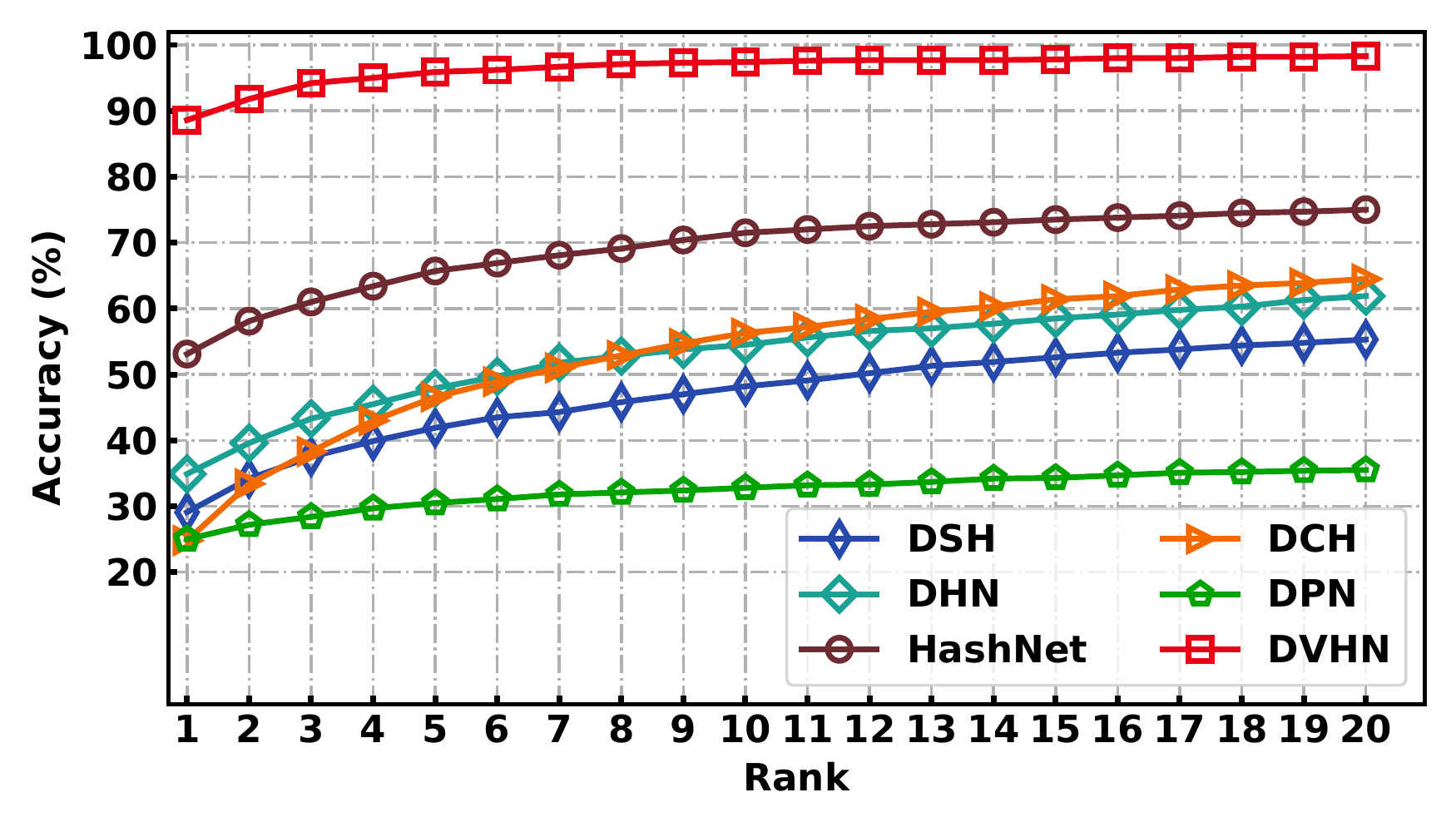}}
\caption{The Cumulative Matching Characteristics (CMC) results against the State-of-the-art methods on three datasets across 256, 512, 1024, 2048 hashing bits} 
\label{fig:cmc}  
\end{figure*}

We further  provide the top-K accuracy results (\textbf{CMC} curves)  of different hash bit lengths on  \textbf{VehicleID}~\textit{(800)},~\textbf{VehicleID}~(1600) and \textbf{VeRi} in Fig.~\ref{fig:cmc} . \par
\textbf{VehicleID}~\textit{(800)}. As depicted in Fig.~\ref{fig:cmc}, we compare the \textbf{CMC} results of our model \textbf{DVHN} with other state-of-the-art deep supervised hash methods across four different hashing bit lengths. Note that we do not include two unsupervised counterparts into comparisons considering their unsatisfactory performance in this domain. Evidently,~\textbf{DVHN} achieves notable results especially from \textbf{Rank@1} to \textbf{Rank@5}. Specifically, we achieve 11.52\%,~9.61\%,~10.75\% and 11.05\% performance gains against \textbf{DPN} in 256, 512, 1024 and 2048 hash bits. It is also easy to note that \textbf{HashNet} can achieve comparable performance from \textbf{Rank@6} to \textbf{Rank@20} when hash bit length equals 256 and 512. However, in terms of vehicle re-identification task, it is of crucial importance that the relative matching images appear in the front of the retrieval list. In terms of \textbf{mAP}, as depicted in the above part of Fig.~\ref{fig:cmc}, \textbf{DVHN} beats all the other competing methods with notable margins. It is also easy to spot that with the number of bits increasing, our proposed model achieves growing \textbf{mAP} scores from 71.61\% to 76.81\%. The performance gain could be attributed to the fact that the model capture more information with a comparatively larger hash bit length. \\

\textbf{VehicleID}~\textit{(1600)}. For test performance on this scenario, as demonstrated in Fig.~\ref{fig:cmc}, a similar trend is spotted for both \textbf{CMC} and \textbf{mAP}. We achieve 57.06\%, 60.41\%, 63.26\% and 64.11\% of \textbf{Rank@1} for 256, 512, 1024 and 2048 hash bits respectively. In regard to \textbf{mAP}, we outperform all the candidate methods by large margins in all testing scenarios. Note that, when compared with performance on \textbf{VehicleID}\textit{(800)}, the model experiences moderate performance declines for all hash bits. For instance, the \textbf{Rank@1} accuracy drops by over 4\% for hash bit 256 while the \textbf{mAP} drops by near 5\%. The degraded performance is mainly due to the introduced difficulty of larger gallery set. \par
\textbf{VeRi}. As is manifested in Fig.~\ref{fig:cmc}, \textbf{DVHN} exhibits consistent advantages against all the competing method across all the evaluation metrics. Specifically, as shown in Fig.~\ref{fig:cmc}, \textbf{DVHN} outperforms \textbf{HashNet} by large margins, consistently. We achieve 80.10\%, 85.81\%, 87.01\% and 88.62\% of \textbf{Rank@1} for 256, 512, 1024 and 2048 bits, beating \textbf{HashNet} by 30.93\%, 33.85\%, 33.73\% and 35.46\%. Similar performance gains could be spotted for other hash bit lengths. As for \textbf{mAP}, the performance gap between our model and other methods is even more pronounced. We outperform the second best model by 32.77\% when hash bit length equals 2048. It is obvious to note that in Fig.~\ref{fig:cmc}, the curve of \textit{DVHN} which is colored red consistently levitates above other competing methods with notable margins. 
The superiority of our method could mainly be attributed to the joint design of the similarity-preserving learning and the discrete hashing learning.

\begin{figure}
	\centering
	\subfigure[The convergence curve of \textbf{DVHN} on 256 (left) and 512 (right) bits.]{
		\begin{minipage}[b]{0.5\textwidth}
			\includegraphics[width=0.45\textwidth]{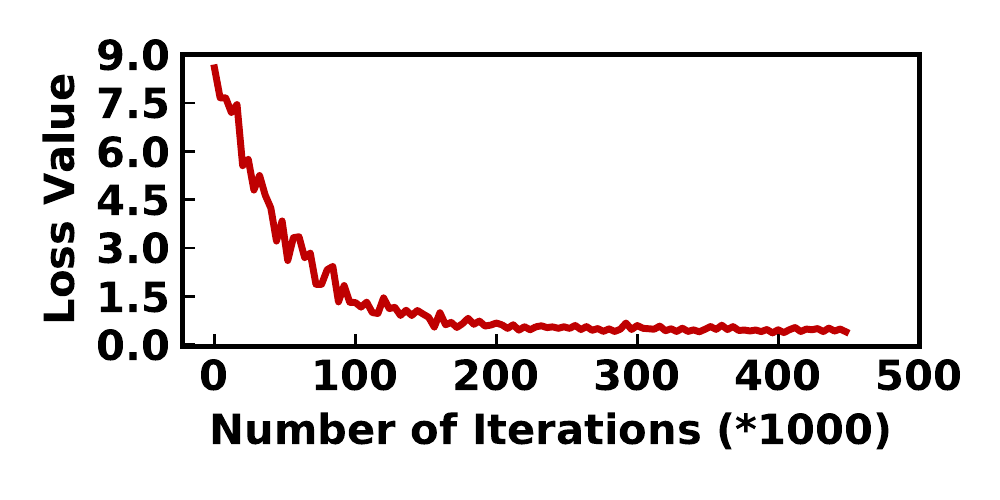} 
			\includegraphics[width=0.45\textwidth]{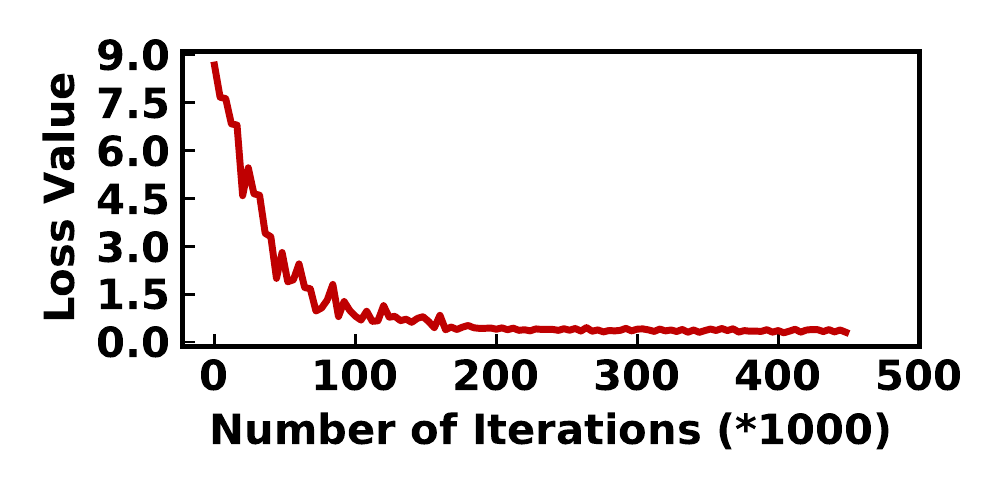}
		\end{minipage}
		}
    	\subfigure[The convergence curve of \textbf{DVHN} on 1024 (left) and 2048 (right) bits.]{
    		\begin{minipage}[b]{0.5\textwidth}
   		 	\includegraphics[width=0.45\textwidth]{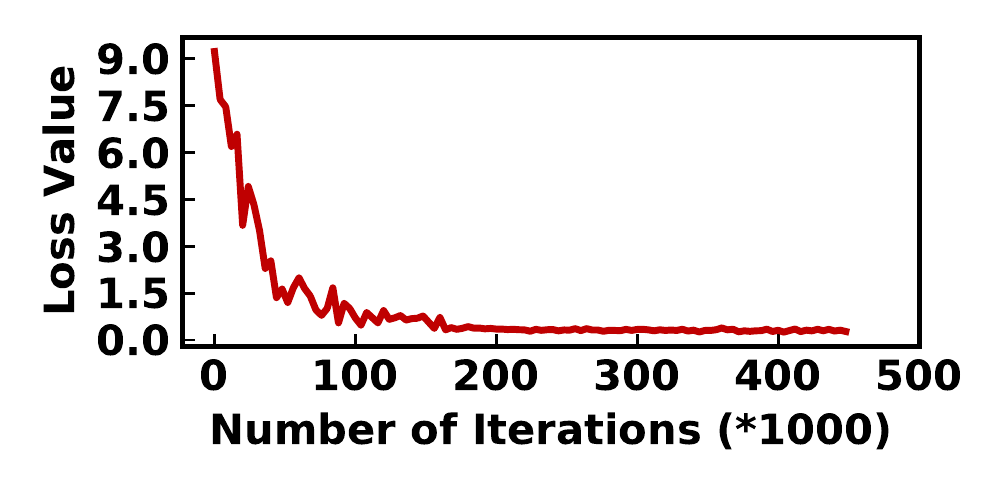}
		 	\includegraphics[width=0.45\textwidth]{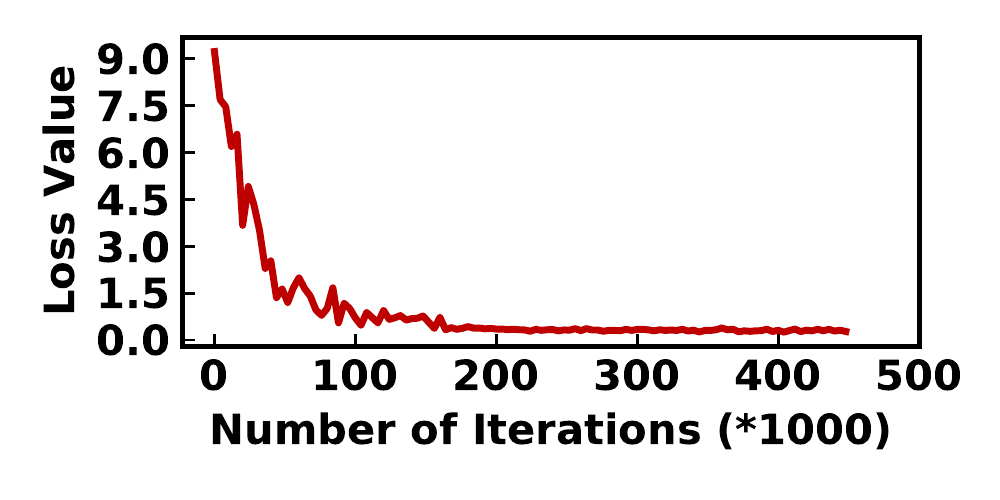}
    		\end{minipage}
	
    	}
	\caption{The Convergence Analysis of \textbf{DVHN} of 256, 512, 1024 and 2048 bits.}
	\label{fig:tloss}
\end{figure}

\begin{figure}
	\centering
	\subfigure[Query Time on VehicleID (1600)]{
	\begin{minipage}[t]{1\linewidth}
		\centering
		\includegraphics[width=3.2in]{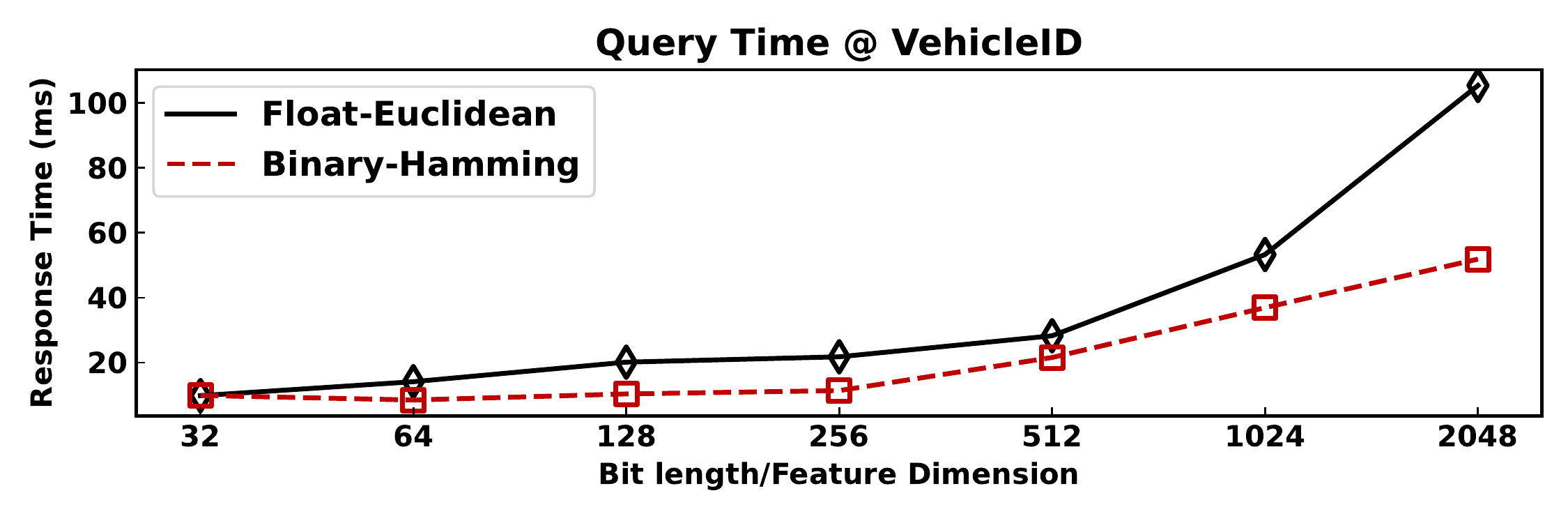}
		\label{fig:vehiclespeed}
	\end{minipage}}
	\\
	\subfigure[Storage Cost on VehicleID (1600)]{
	\begin{minipage}[t]{1\linewidth}
		\centering
		\includegraphics[width=3.2in]{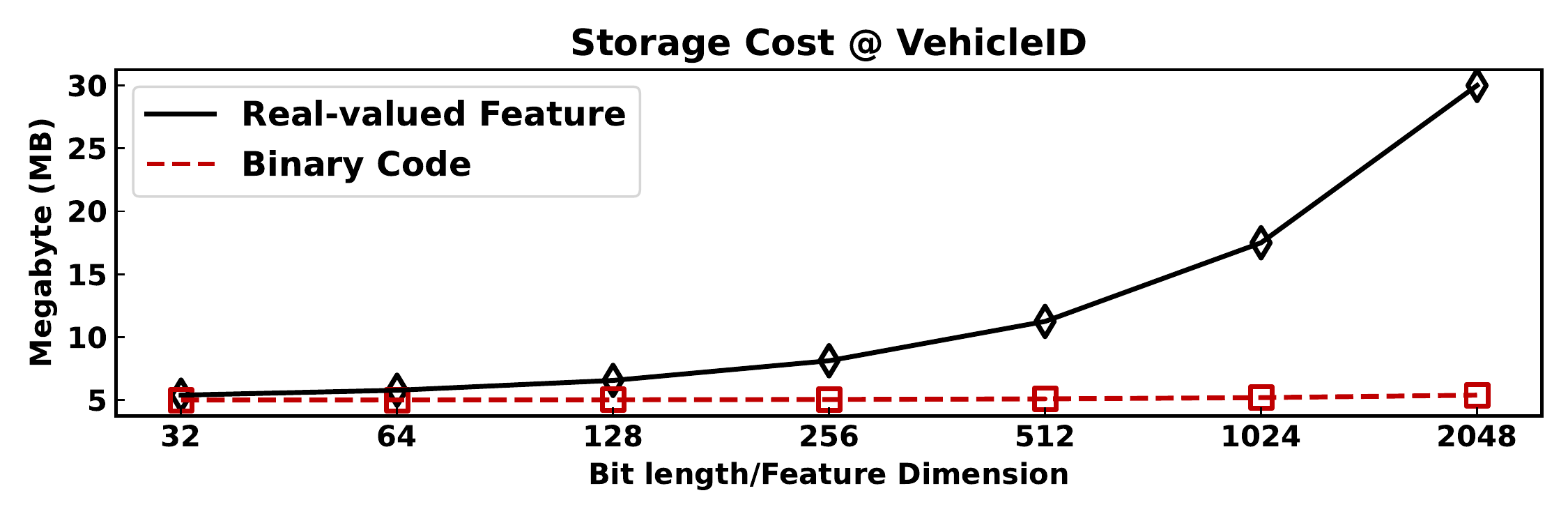}
		\label{fig:vehiclestorage}
	\end{minipage}}
	\\
	\subfigure[Query Time on VeRi]{
	\begin{minipage}[t]{1\linewidth}
		\centering
		\includegraphics[width=3.2in]{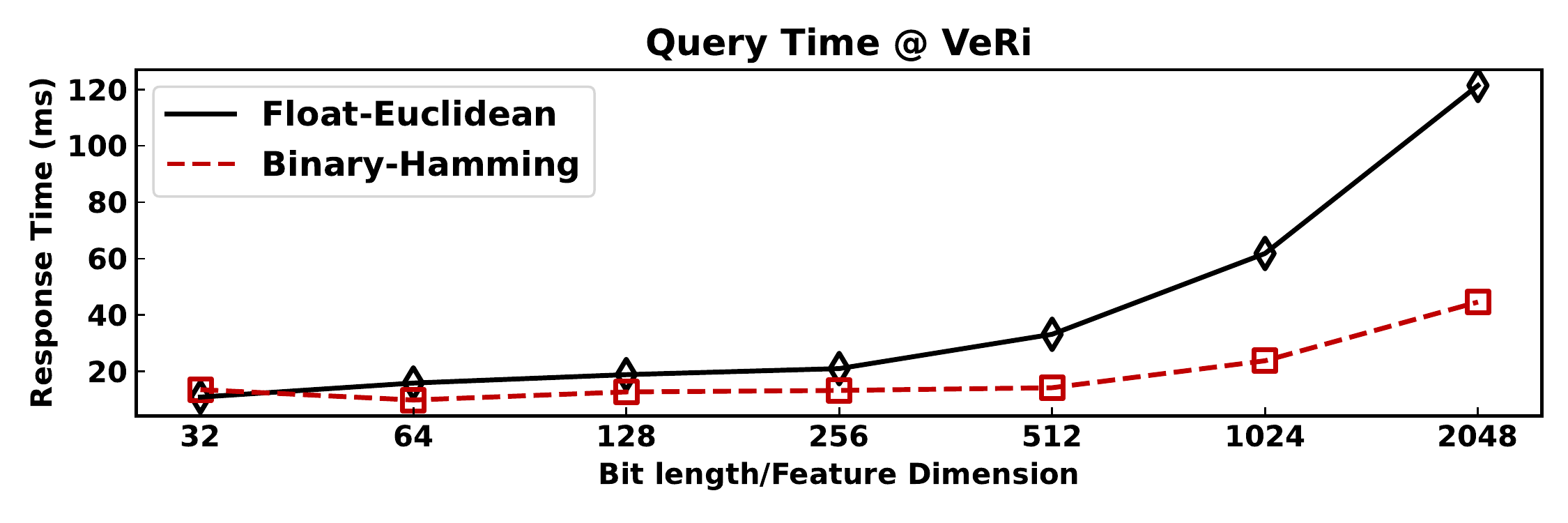}
		\label{fig:verispeed}
	\end{minipage}}
	\\
	\subfigure[Storage Cost on VeRi]{
	\begin{minipage}[t]{1\linewidth}
		\centering
		\includegraphics[width=3.2in]{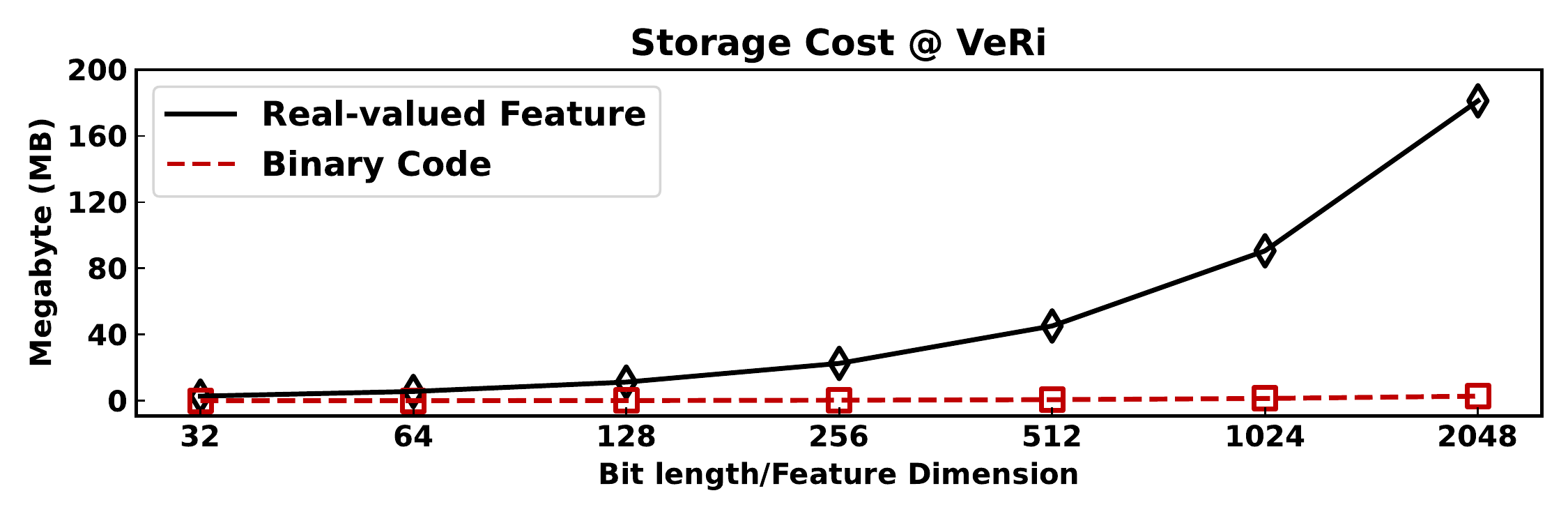}
		\label{fig:veristorage}
	\end{minipage}}
	\caption{Comparison of the total query response time and storage cost of }
	\label{fig:qnscost}
\end{figure}

\begin{figure}
	\centering
	\subfigure[The CMC results of different variants on \textbf{VeRi} dataset]{
	\begin{minipage}[t]{1\linewidth} 
		\centering
		\includegraphics[width=2.8in]{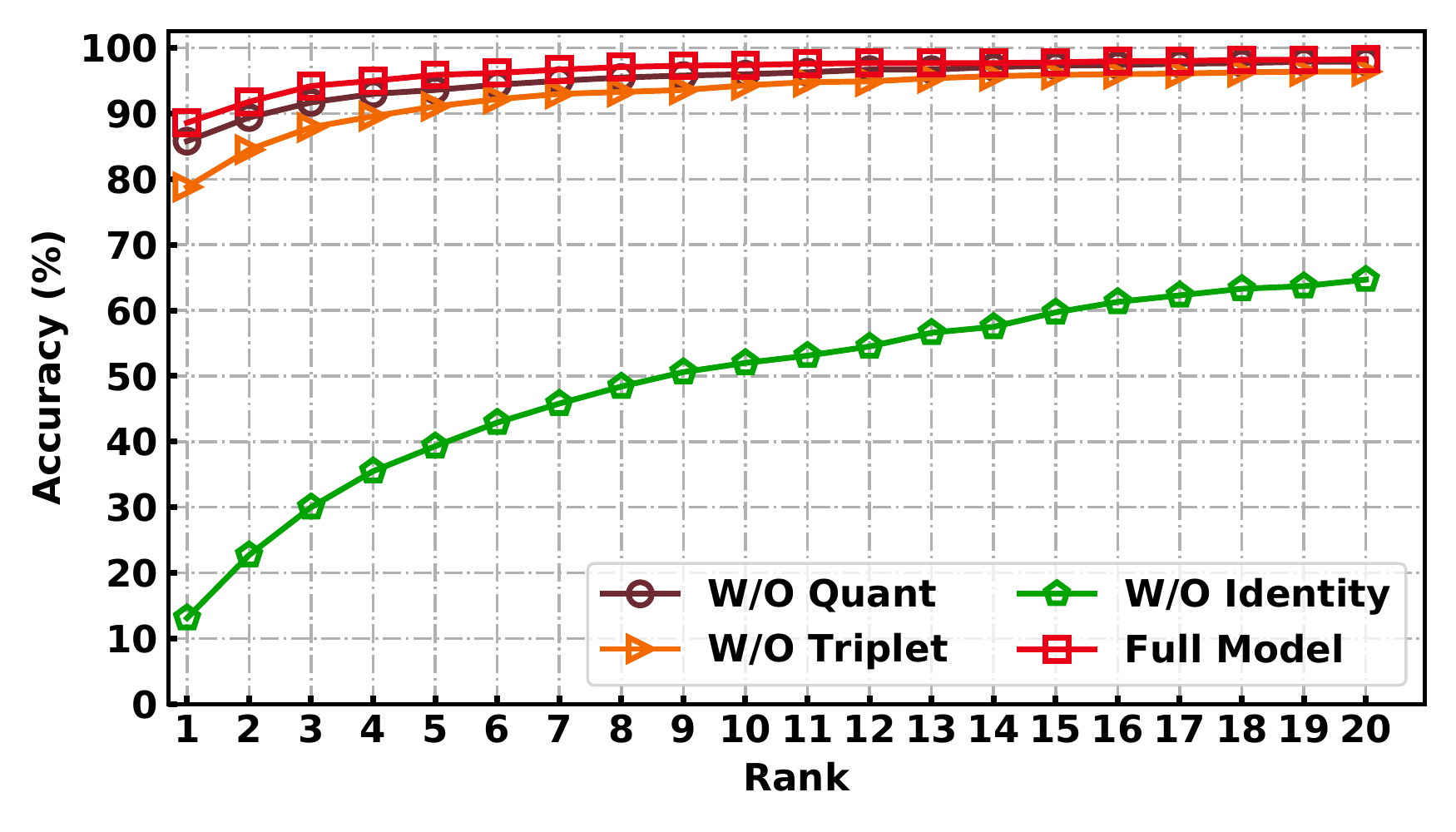}
		\label{fig:abcmcveri}
	\end{minipage}}
	\\
	\subfigure[The CMC results of different variants on \textbf{VehicleID (800))} dataset]{
	\begin{minipage}[t]{1\linewidth}
		\centering
		\includegraphics[width=2.8in]{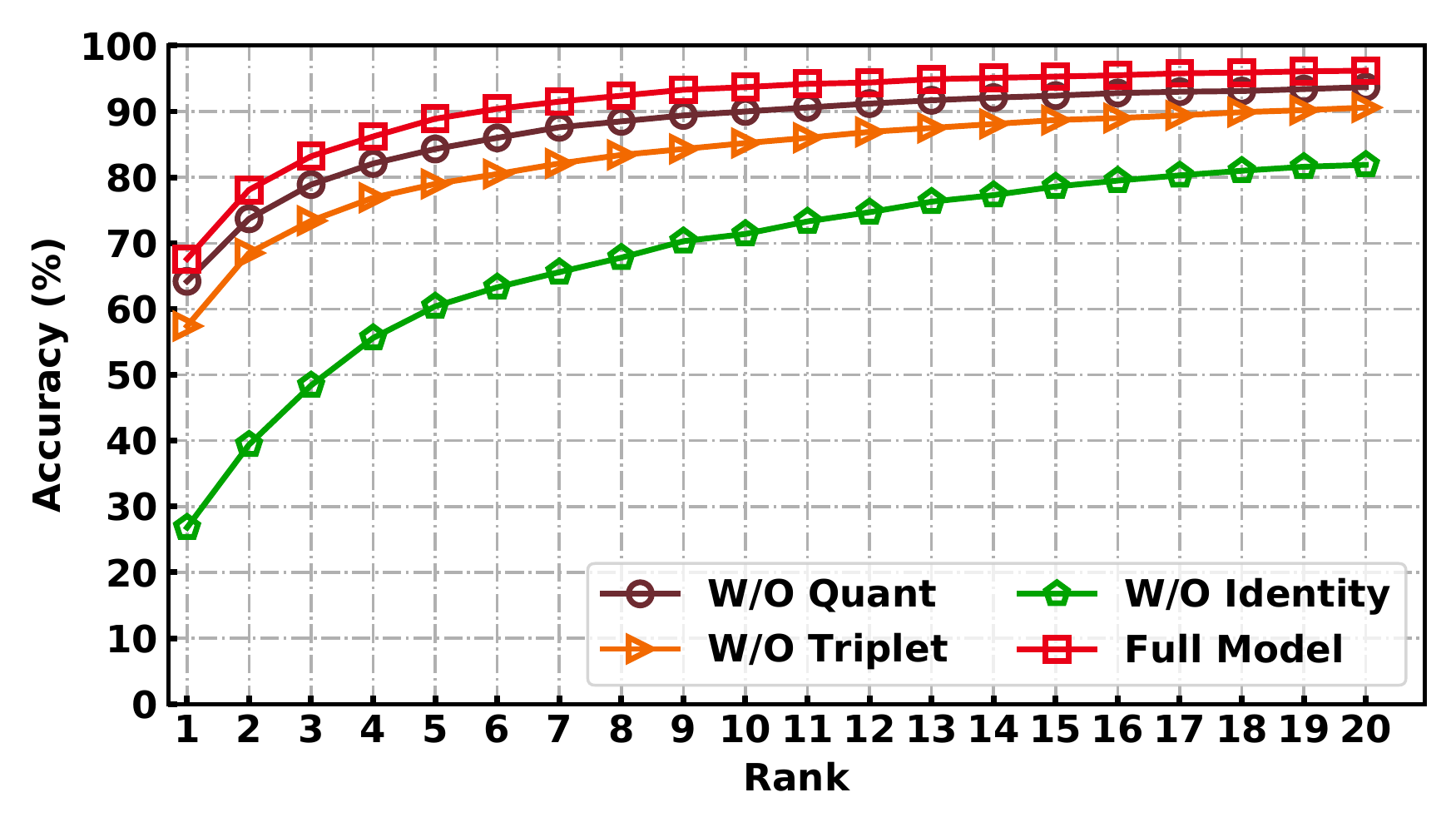}
		\label{fig:abcmcvid}
	\end{minipage}}
	\\
	\subfigure[The mAP results of different variants on both \textbf{VeRi} and \textbf{VehicleID}(800) datasets]{
	\begin{minipage}[t]{1\linewidth}
		\centering
		\includegraphics[width=3.2in]{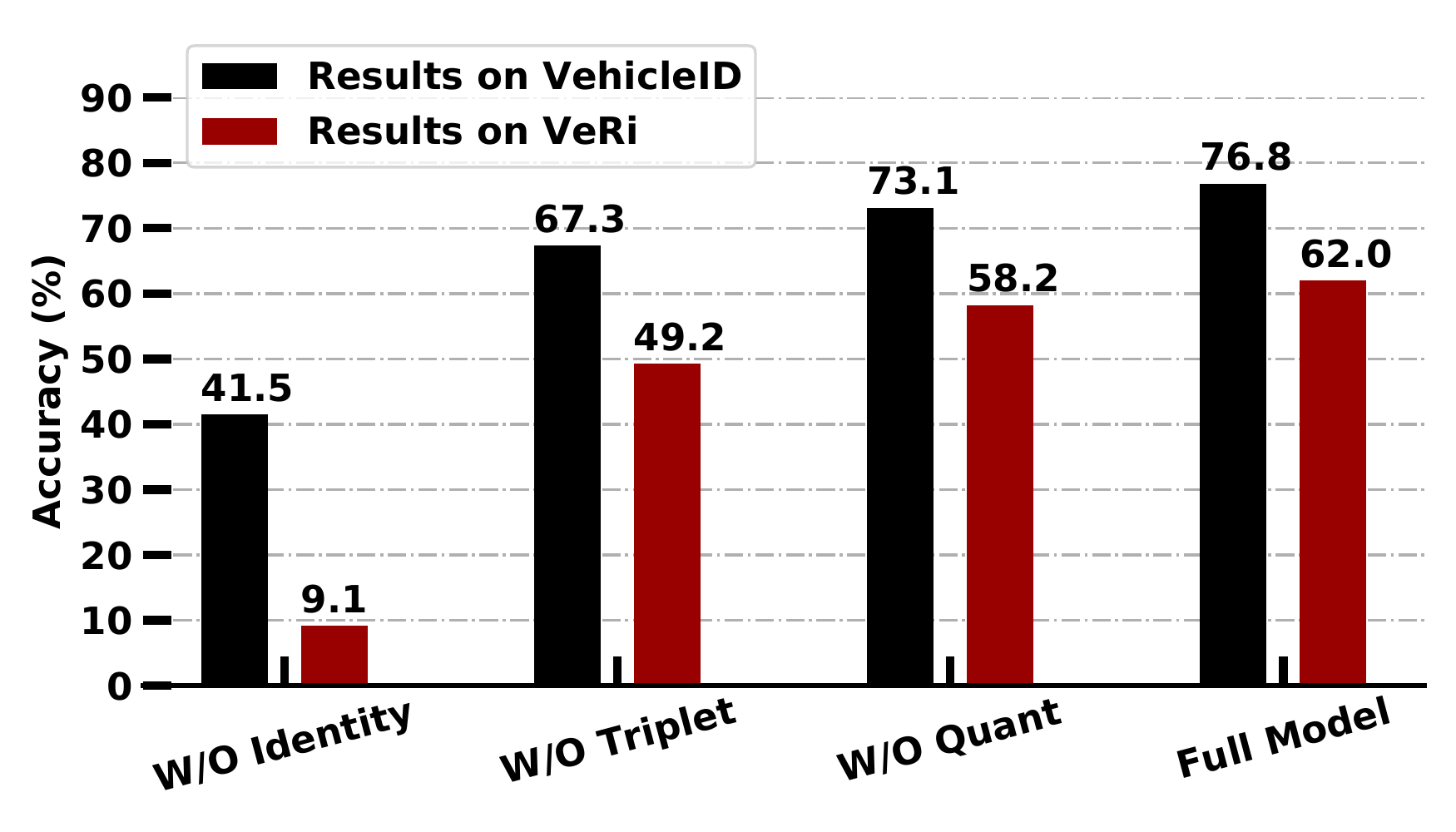}
		\label{fig:abmap}
	\end{minipage}}
	\caption{The CMC and mAP Results of DVHN and Its Three Variants}
	\label{fig:ablation}
\end{figure}

\subsection{Empirical Analysis}
First, we plot the convergence curves for our model across four hashing bits in Fig.~\ref{fig:tloss}. \par
Second, we conduct a detailed ablation study to demonstrate the effectiveness of each component of our proposed \textbf{DVHN}. To effectively evaluate the importance of \textbf{Identity Loss} ($L_{Identity}$), triplet-based hash loss ($L_{Triplet}$), and the proposed discrete learning loss ($L_{Quant}$), we empirically test the performance of our model on \textbf{VehicleID} \textit{(800)} and \textbf{VeRi} when stripped of each component. We denote the model of all the component as \textbf{Full Model}, the model without triplet loss as \textbf{W/O Triplet}, the model without \textbf{Identity Loss} as \textbf{W/O Identity} and, finally, the model without the discrete hashing learning as \textbf{ W/O Quant}. Note that all the experiments were conducted on 2048 hash bits. As demonstrated in Fig.~\ref{fig:abcmcveri} and Fig.~\ref{fig:abcmcvid}, when the discrete learning loss  is moved from the model, we experience slight performance declines, 3.46\% and 3.5\% for \textbf{Rank@1} and \textbf{mAP} on \textbf{VehicleID (800)}, and 2.80\% and 3.77\% for \textbf{Rank@1} and \textbf{mAP} on \textbf{VeRi}. This evidences the validity of the discrete hashing learning module for the binary \textit{Hamming} hashing codes to preserve the class information. When deprived of the identity loss, the model experience conspicuous performance decreases. On \textbf{VehicleID (800)}, the \textbf{Rank@1} accuracy drops by 40.84\% while the \textbf{mAP} score drops by 35.31\%. On \textbf{VeRi}, the performance deterioration is even more severe, 75.11\% for \textbf{Rank@1} and 52.84\% for \textbf{mAP}. The declines have demonstrated the importance of the identity loss on learning more robust and discriminative features. Finally, when looking at the performance of \textbf{W/O Triplet}, we could also spot notable performance declines against the \textbf{Full Model}. \textbf{W/O Triplet} achieves 57.47\% of \textbf{Rank@1} accuracy and 67.32\% \textbf{mAP}, 10.19\% and 9.45\% smaller than its full model counterpart on \textbf{VehicleID (800)}. On \textbf{VeRi}, a similar trend can also be spotted when the model is depried of $L_{Triplet}$. \textbf{W/O Triplet} achieves 78.84\% of \textbf{Rank@1} and 49.23\% of \textbf{mAP}, 9.78\% and 12.79\% lower than the \textbf{Full Model} counterpart.The deterioration is largely due to the effectiveness of batch-hard triplet loss in tackling the vehicle re-identification problem, where the intra-class variation is sometimes even larger than the inter-class variation. Finally, as elaborated above, all the proposed components are essential to the final performance of the proposed final model.

\begin{figure*}
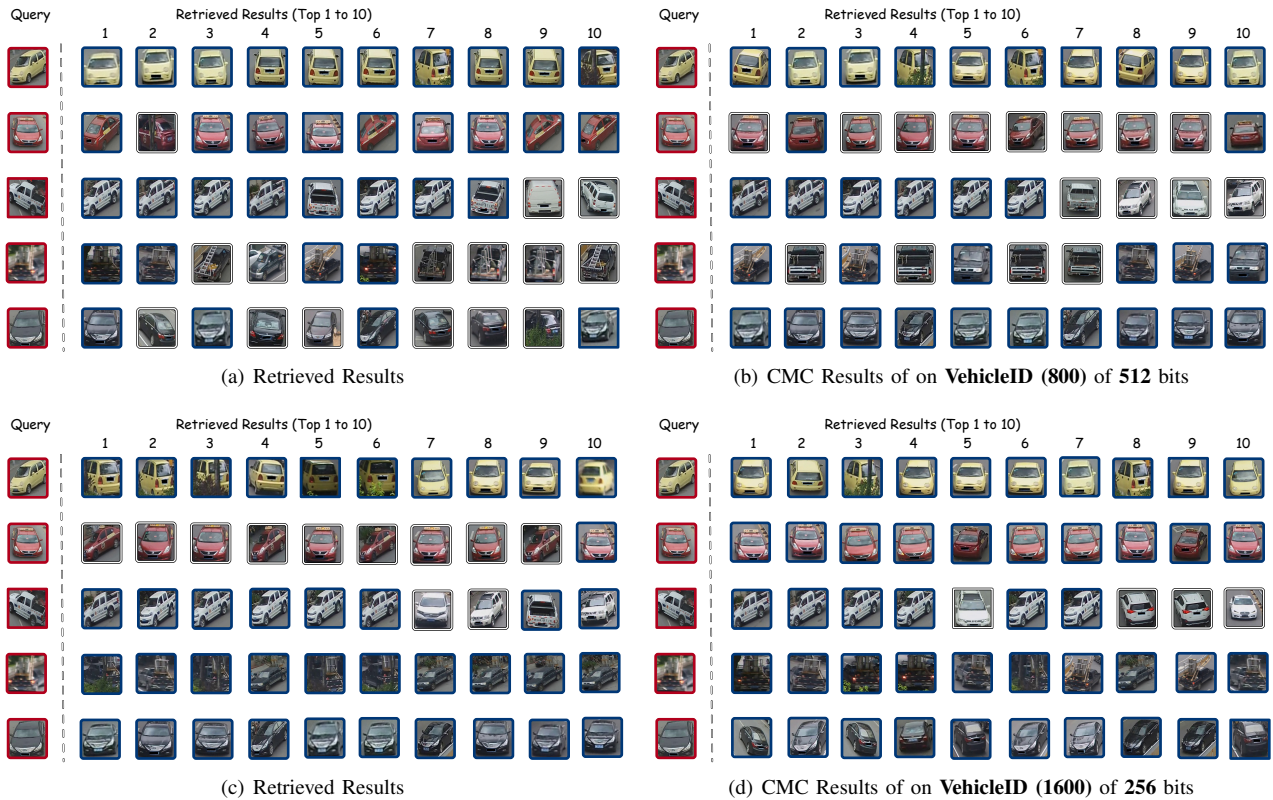

\centering
\subfigure[Retrieved Results] 
{\includegraphics[width=8.5cm]{images/256_retrieved_results.pdf}}
\subfigure[CMC Results of on \textbf{VehicleID (800)} of \textbf{512} bits]{\includegraphics[width=8.5cm]{images/512_retrieved_results.pdf}}
\\
\subfigure[Retrieved Results] 
{\includegraphics[width=8.5cm]{images/1024_retrieved_results.pdf}} 
\subfigure[CMC Results of on \textbf{VehicleID (1600)} of \textbf{256} bits]{\includegraphics[width=8.5cm]{images/2048_retrieved_results.pdf}}

\caption{The Visualization of the Top 10 Retrieved Results of Our Framework on 256, 512, 1024, 2048 Hashing Bits. } 
\label{fig:visual}  
\end{figure*}

\subsection{Retrieval Efficiency Comparison}
The time cost of a \textbf{Vehicle ReID} system generally is comprised of two separate parts: feature extraction cost and similarity calculation cost. As is pointed out by \cite{chen2020deep}, the cost of the feature extraction actually heavily depends on the backbone network used. Thus, we focus on comparing the large-scale similarity calculation time costs of our method and the real-valued vehicle re-identification counterparts. Meanwhile, we also compare the storage costs of our \textit{Hamming} hashing re-identification framework with the real-valued ones. \par
Specifically, we illustrate the results on \textbf{VehicleID (1600)} and \textbf{VeRi} in Fig.~\ref{fig:qnscost}, which have (11777,~1600), (1678,~11579) (query, gallery) images, respectively. In Fig.~\ref{fig:vehiclespeed} and Fig.~\ref{fig:verispeed}, the 'Float-Euclidean' denotes the query time for Euclidean distance based real-valued vehicle re-identification methods while 'Binary-Hamming' denotes the query time for our binary hashing method with \textit{Hamming} distance calculation. For Fig.~\ref{fig:vehiclestorage} and Fig.~\ref{fig:veristorage},~it demonstrates the comparison of storage cost on \textbf{VehicleID (1600)} and \textbf{VeRi} for real-valued methods and our binary \textit{Hamming} codes based method. It is obvious that binary codes could save tremendous memory compared with its real-valued counterparts.

\subsection{Qualitative Analysis of the Retrieved Results}

In Fig.~\ref{fig:visual}, we visualize the retrieval results for different hash bit length of our model in \textbf{VeRi}. Note that, to demonstrate the retrieval efficiency, we take the original gallery set as a query set so that for each query image, there exist several images in the gallery set. In the picture above, we adopt the same query images for four testing scenarios where the hash bit length equals 256, 512, 1024 and 2048, respectively.  The right part visualizes the retrieval list for the query image. The image enclosed by a red box means it is from the same vehicle as the query image. Generally, for a re-identification model, it would be desirable for the matching images to appear in the front of the retrieval list. 
\section{Conclusion}
In this paper, we propose \textbf{DVHN}, a deep hashing based framework for efficient large-scale vehicle re-identification. Contrary to conventional method which adopt real-valued feature vectors to perform re-identification task, we adopt a \textit{Hamming} hashing-based scheme which saves tremendous memory and is much faster in terms of retrieval speed and thus, could be applied in a large-scale re-identification scenario. Specifically, for similarity-preserving learning, we adopt a triplet loss and an on-the-fly hard triplet generation module. Meanwhile, we further add an identity loss to learn more robust and discriminative features. For binary code learning, we propose a discrete hashing learning scheme which is based on the assumption that the binary codes should be ideal for classification. To overcome the vanishing gradient problem (the \textit{sign} function for generating binary codes is not differentiable), we adopt an alternative optimization method for optimizing the overall framework. We conduct extensive experiments and the results have demonstrated that our framework could generate discriminative binary codes for efficient vehicle re-identification, surpassing the stet-of-the-art hashing methods with large margins.

\bibliographystyle{IEEEtran}
\bibliography{main}

\vfill


\end{document}